\documentclass{article} 
\usepackage{iclr2027_conference,times}

\usepackage{hyperref}
\usepackage{amsmath,amsfonts,amssymb}
\hypersetup{
    colorlinks,
    linkcolor={red!50!black},
    citecolor={blue!50!black},
    urlcolor={blue!80!black}
}
\usepackage{url}
\usepackage{algorithm}
\usepackage[noend]{algpseudocode}
\usepackage{booktabs}
\usepackage{multirow}
\usepackage{graphicx}
\usepackage{subcaption}

\newcommand{\diff}{\mathop{}\!\mathrm{d}}

\title{Interacting Particle Guidance for Sampling\\ Reward-tilted Generative Priors}

\author{
Adhithyan Kalaivanan$^1$ \quad Zheng Zhao$^1$ \quad Jens Sjölund$^2$ \quad Fredrik Lindsten$^{1}$ \\
$^1$Linköping University, Sweden \quad $^2$Uppsala University, Sweden\\
\texttt{\{adhithyan.kalaivanan,zheng.zhao,fredrik.lindsten\}@liu.se}\\
\texttt{jens.sjolund@it.uu.se}\\
}

\iclrfinalcopy 
\begin{document}

\maketitle

\begin{abstract}

Inference-time steering adapts pretrained diffusion and flow-based models to new tasks, e.g., to generate samples from a conditional distribution or samples with desired properties, without retraining. 
This can be formalized as sampling from a reward-tilted generative prior.
As exact sampling from this distribution is intractable, guidance-based methods rely on approximations producing biased samples, and sequential Monte Carlo (SMC) methods correct for this bias using importance weights.
However, while exact in the large particle limit, SMC suffers from weight degeneracy and particle collapse in practice.
We propose interacting particle guidance (IPG), which replaces reweighting with transport.
The particles interact through an additional drift, derived from the Feynman--Kac PDE to cancel the reweighting term, and remain unweighted.
Choosing the drift in a reproducing kernel Hilbert space yields a closed-form solution that is cheap to compute, with negligible overhead compared to SMC.
We demonstrate the method on Gaussian mixtures with known posteriors, and on high-dimensional image inpainting and protein structure inference tasks.

\end{abstract}

\section{Introduction}

Generative models based on dynamic measure transport, such as diffusion \citep{ho2020denoising, song2020score} and flow-based models \citep{lipman2022flow, liu2022flow, albergo2025stochastic}, are the state of the art approaches in a wide range of domains.
Often, beyond unconditionally sampling from the data distribution, we may want to generate samples with specific properties or conditionally on some observed data.
A convenient way to do so is to steer the generation process at test time, which allows reusing pretrained models without task-specific retraining.
For a given reward function or observation likelihood, this can be formalized as sampling from an exponentially reward-tilted generative prior.

For differentiable reward functions, a heuristic approach is to include the reward gradient as a guidance term during generation \citep{chung2023diffusion, boys2024tweedie}.
However, the modified process does not offer any guarantees of sampling from the reward-tilted prior.
A principled approach is to correct for this using sequential Monte Carlo (SMC) methods \citep{naesseth2019elements}, in which particles with self-normalized importance weights provide a weighted empirical distribution that converges to the target distribution in the large particle limit.
In practice, however, when the prior is given by a generative model, the number of particles is limited by GPU memory, leading to pronounced weight degeneracy and particle collapse \citep{wu2023practical, singhal2025a, skreta25a, kelvinius25b}.
As a result, in practical implementations of SMC-based methods with generative priors and a moderate number of particles, reweighting and resampling are often interpreted as local best-of-N search strategies \citep[Appendix A.6]{millard2026particleguided}.

To mitigate the weight degeneracy, we analyze the dynamics of the intermediate target densities in the continuous time limit, given by the Feynman--Kac PDE \citep{skreta25a}.
By introducing an additional transport term that cancels the reweighting term, we aim to derive unweighted particle updates. 
Similar approaches that minimize the reweighting term have been explored for sampling unnormalized densities, where the added transport is parameterized by a neural network and learned over simulated trajectories \citep{albergo25a}, or at each intermediate step using the current particles \citep{arbel2021annealed}.
To preserve the training-free nature of inference-time steering, the drift must be parameterized such that it admits a closed-form solution.
Prior work has studied linear parameterizations with predefined bases \citep{ren2026driftlite}, but as these do not fully cancel the reweighting term, they still require periodic resampling to maintain a high effective sample size.
Building on Stein transport \citep{nusken2024stein}, we show that choosing the drift in a reproducing kernel Hilbert space (RKHS) is surprisingly effective and enables sampling with an unweighted interacting particle system.
The drift is available in closed form, and is cheap to compute for the moderate number of particles used with generative priors.

\paragraph{Contributions.} 
(\textit{i}) We propose interacting particle guidance (IPG), which augments guidance-based methods with an interacting drift estimated to cancel the reweighting term in the Feynman--Kac PDE, enabling sampling from a reward-tilted generative prior with an \emph{unweighted} interacting particle system. The drift admits a closed-form solution, with negligible overhead to compute.
(\textit{ii}) We empirically show that IPG avoids particle collapse and outperforms prior methods, on Gaussian mixtures with known posteriors, as well as on high-dimensional image inpainting and protein structure inference with flow-based generative priors. 

\section{Preliminaries}

\subsection{Diffusion and flows}
Consider a generative model that transports samples $\{X^i_0\}_{i=1}^N$ from a simple base distribution $q_0$ at time $t=0$ to the data distribution $q_1$ at $t=1$ along a learned vector field $v_t(x)$, by solving the ordinary differential equation (ODE)
\begin{equation}
    \diff X_t^i = v_t(X_t^i) \diff t, \quad X_0^i \sim q_0.
\end{equation}

Both flow-based models \citep{lipman2022flow, liu2022flow, albergo2025stochastic} and diffusion models using the probability flow ODE \citep{song2020score} can be viewed as instances of this model class.
The intermediate marginal densities $q_t$ satisfy the continuity equation,
\begin{equation} \label{eq:continuity}
    \partial_t q_t(x) = - \nabla \cdot (v_t(x) q_t(x)).
\end{equation}
In the next section, we derive an analogous PDE for reward-tilted intermediate densities.

\subsection{Reward tilting} \label{sec:rewardtilting}
Generating samples with specific properties or conditionally on observed data can be formulated as sampling from a reward-tilted prior $p_1(x) \propto q_1(x) \exp(R(x))$ for a reward function $R(x)$.
We can do this by evolving particles along an interpolating path of densities $p_t(x) \propto q_t(x) \exp(r(x, t))$, with $r(x, 0) = 0$ and $r(x, 1) = R(x)$, so that $p_0 = q_0$ is the base distribution.
We show in Appendix~\ref{sec:rtfkp} that $p_t$ satisfies
\begin{equation} \label{eq:partialpt}
    \partial_t p_t(x) = - \nabla \cdot (v_t(x) p_t(x)) + p_t(x) (g_t(x) - \mathbb{E}_{p_t}[g_t]),
\end{equation}
where $g_t(x) = \partial_t r(x, t) + \langle v_t(x), \nabla r(x, t) \rangle$.

Equation~\eqref{eq:partialpt} takes the form of a Feynman--Kac PDE \citep{skreta25a}, and can be simulated with a weighted particle system $\{(X_t^i, w_t^i)\}_{i=1}^N$ whose weighted empirical distribution approximates $p_t$,
\begin{equation}
    \diff X_t^i = v_t(X_t^i) \diff t, \quad \diff \omega_t^i = g_t(X_t^i) \diff t,
\end{equation}
where the log-weights $\omega_t^i$ are initialized at zero, and $w_t^i = \exp(\omega_t^i)/\sum_{k} \exp(\omega_t^k)$ are the self-normalized importance weights. 
However, this approach is equivalent to importance sampling with the prior $q_1$ as the proposal, and performs poorly for small values of $N$.

\paragraph{Adding drift-diffusion.} A popular strategy to let the reward influence the particle updates is to add and subtract $\sigma_t \Delta p_t(x)$, for $\sigma_t \in \mathbb{R}_{>0}$, in equation~\eqref{eq:partialpt}, and write
\begin{equation}
    -\sigma_t \Delta p_t(x) = -\nabla \cdot ((\sigma_t \nabla \log p_t(x)) p_t(x)).
\end{equation}
The resulting modified PDE for $p_t$ is given by
\begin{equation}
    \partial_t p_t(x) = - \nabla \cdot ((v_t(x) + \sigma_t \nabla \log p_t(x))p_t(x)) + \sigma_t \Delta p_t(x) + p_t(x) (g_t(x) - \mathbb{E}_{p_t}[g_t]),
\end{equation}
which, as before, can be simulated using the weighted stochastic differential equations (SDEs),
\begin{equation} \label{eq:sdetilt}
    \diff X_t^i = (v_t(X_t^i) + \sigma_t \nabla \log p_t(X_t^i)) \diff t + \sqrt{2\sigma_t} \diff W_t, \quad \diff \omega_t^i = g_t(X_t^i) \diff t.
\end{equation}

Here, $\nabla \log p_t(x) = \nabla \log q_t(x) + \nabla r(x, t)$, where the score $\nabla \log q_t(x)$ is directly estimated in diffusion models, or computed from $v_t$ in flow-based models with a Gaussian base distribution \citep{albergo2025stochastic}.
For the choice $r(x, t) = \beta_t R(x)$, equation~\eqref{eq:sdetilt} recovers the reward-tilted SDE of \citet{skreta25a}, where the weighted SDEs are simulated using SMC, resampling the particles whenever the weights become too skewed.
However, in high dimensions and with the few particles affordable with generative priors, the weights degenerate and resampling leads to particle collapse \citep{naesseth2019elements}.

\section{Method} \label{sec:method}

The Langevin terms in equation~\eqref{eq:sdetilt} influence the particle updates, but leave the weights unchanged.
Our methodology is based on compensating for the weights by an additional drift, as detailed below.

\paragraph{Adding drift-reweighting.} For any vector field $u_t$, we add and subtract $\nabla \cdot(u_t(x) p_t(x))$ in equation~\eqref{eq:partialpt}, and write
\begin{equation}
    \nabla \cdot(u_t(x) p_t(x)) = p_t(x) (S_{p_t} u_t(x)),
\end{equation}
where $S_{p_t}u_t = \nabla \cdot u_t + \langle u_t, \nabla \log p_t \rangle$ is the Stein operator.
The modified PDE for $p_t$ is
\begin{equation}
    \partial_t p_t(x) = - \nabla \cdot ((v_t(x) + u_t(x)) p_t(x)) + p_t(x) (S_{p_t}u_t(x) + g_t(x) - \mathbb{E}_{p_t}[g_t]).
\end{equation}
As $\mathbb{E}_{p_t}[S_{p_t} u_t (x)] = 0$ by Stein's identity, $p_t$ can be approximated with the weighted particle system
\begin{equation}
    \diff X_t^i = (v_t(X_t^i) + u_t(X_t^i)) \diff t, \quad \diff \omega_t^i = (S_{p_t}u_t(X_t^i) + g_t(X_t^i)) \diff t.
\end{equation}
As before, the Langevin terms can be added without changing the weight updates, since they leave $p_t$ invariant.

We note that the drift $u_t$ can be chosen to reduce the variance of the particle weights, and in particular, if $u_t$ solves the PDE
\begin{equation} \label{eq:steinpde}
    S_{p_t}u_t(x) + g_t(x) - \mathbb{E}_{p_t}[g_t] = 0,
\end{equation}
then the log-weights remain constant and $p_t$ is approximated by \emph{unweighted} particles. We refer to such $u_t$ as a corrective drift.

\subsection{Interacting particle guidance} \label{sec:ipg}

As solving the PDE~\eqref{eq:steinpde} exactly is intractable, we minimize the squared residual locally at each time.
Following Stein transport \citep{nusken2024stein}, choosing $u_t \in \mathcal{H}_k^d$, where $\mathcal{H}_k$ is the reproducing kernel Hilbert space (RKHS) for a kernel $k$ and $\mathcal{H}_k^d$ is its $d$-fold Cartesian product, and adding Tikhonov regularization with $\lambda > 0$, yields the objective
\begin{equation} \label{eq:ipsobjective}
    \mathcal{L}(u_t) = \mathbb{E}_{p_t}[(S_{p_t}u_t(x) + g_t(x) - \mathbb{E}_{p_t}[g_t])^2] + \lambda \|u_t\|_{\mathcal{H}_k^d}^2.
\end{equation}
Approximating $\mathbb{E}_{p_t}[\cdot]$ with the empirical distribution given by the particles $\{X_t^i\}_{i=1}^N$, the unique minimizer is
\begin{equation} \label{eq:rkhsdrift}
    u_t(x) = \frac{1}{N}\sum_{j=1}^N \phi_j [k(x, X_t^j) \nabla \log p_t(X_t^j) + \nabla_{X_t^j}k(x, X_t^j)], \quad \bigg(\frac{1}{N}\xi + \lambda I_N\bigg)\phi = -\mathrm{g}_t,
\end{equation}
where the coefficients $\{\phi_j\}_{j=1}^N$ are obtained by solving the $N \times N$ linear system.
Here, $\xi \in \mathbb{R}^{N \times N}$ is the Gram matrix of the Stein kernel, with $[\xi]_{ij} = S_{p_t}^{X_t^i}S_{p_t}^{X_t^j}k(X_t^i, X_t^j)$ and the superscript denoting the variable the Stein operator acts on, and $[\mathrm{g}_t]_i = g_t(X_t^i) - \frac{1}{N}\sum_{j=1}^N g_t(X_t^j)$. 

As the self-normalized importance weights are invariant to adding a constant to the log-weights of all particles, the weight updates reduce to
\begin{equation}
    \diff \omega_t^i = (S_{p_t}u_t(X_t^i) + g_t(X_t^i) - \frac{1}{N}\sum_{j=1}^N g_t(X_t^j)) \diff t = \bigg(\frac{1}{N}\xi \phi + \mathrm{g}_t\bigg)_i \diff t = - \lambda \phi_i \diff t.
\end{equation}
However, as in Stein transport, the weight updates can be ignored in practice by choosing a small $\lambda$, yielding an unweighted interacting particle system that we refer to as interacting particle guidance (IPG), described in Algorithm~\ref{alg:IPG}.

\paragraph{Using a control variate.} Alternatively, $\mu_t = \mathbb{E}_{p_t}[g_t]$ in objective~\eqref{eq:ipsobjective} can be treated as a free parameter to be jointly minimized.
With the empirical approximation for the outer expectation, the optimal $\mu_t = \frac{1}{N}\sum_{i=1}^N (S_{p_t} u_t(X_t^i) + g_t(X_t^i))$ resembles using a Stein control variate \citep{oates2017control} to estimate $\mathbb{E}_{p_t}[g_t]$.
The unique minimizer is now given by (see Appendix~\ref{sec:ipgcv}),
\begin{equation} \label{eq:rkhscdrift}
    u_t(x) = \frac{1}{N}\sum_{j=1}^N (\Pi \phi)_j [k(x, X_t^j) \nabla \log p_t(X_t^j) + \nabla_{X_t^j}k(x, X_t^j)], \quad \bigg(\frac{1}{N}\Pi\xi\Pi + \lambda I_N\bigg)\phi = -\mathrm{g}_t,
\end{equation}
where $\Pi = I_N - \frac{1}{N}\mathbf{1}\mathbf{1}^\top$ is the centering matrix.
The weight updates once again reduce to,
\begin{equation}
    \diff \omega_t^i = (S_{p_t}u_t(X_t^i) + g_t(X_t^i) - \mu_t) \diff t = \bigg(\frac{1}{N}\Pi\xi\Pi\phi + \mathrm{g}_t\bigg)_i \diff t = - \lambda \phi_i \diff t,
\end{equation}
and can be ignored by choosing a small $\lambda$.
We refer to the resulting method as IPG-CV in Algorithm~\ref{alg:IPG}, emphasizing the use of a Stein control variate.

\paragraph{Connection to Stein transport.} The objective~\eqref{eq:ipsobjective} has the same form as in Stein transport \citep{nusken2024stein}, where particles are transported deterministically from a tractable prior $\pi_0$ to the posterior along $\pi_t \propto \pi_0 \exp(th)$, where $h$ is the log-likelihood, by the corrective drift alone.
In our notation, this corresponds to the case with $v_t = 0$, $\sigma_t=0$, $q_t = \pi_0$ and $r(x, t) = th(x)$.
In IPG, the path is induced by a generative model whose pointwise density is unavailable, particles are transported by the generative drift and guidance, and
the corrective drift only accounts for the remaining reweighting term.

\begin{algorithm}[t]
\caption{Interacting particle guidance (IPG and IPG-CV)}
\label{alg:IPG}
\begin{algorithmic}
\Require $v_t$ and $\nabla \log q_t(x)$ from the generative prior, reward schedule $r(x,t)$, kernel $k$, regularization $\lambda$, noise schedule $\sigma_t$, time steps $\{t_k\}_{k=0}^K$, number of particles $N$
    \State Initialize $X_0^i \sim q_0$ for $i = 1, \ldots, N$
    \For{$t = t_0, \ldots, t_{K-1}$}
    \State $\nabla \log p_t(X_t^i) \leftarrow \nabla \log q_t(X_t^i) + \nabla r(X_t^i, t)$,
    \State Compute $u_t(X_t^i)$ using equations~\eqref{eq:rkhsdrift} for IPG, or~\eqref{eq:rkhscdrift} for IPG-CV,
    \State $X_{t+\Delta t}^i \leftarrow X_t^i + (v_t(X_t^i) + \sigma_t \nabla \log p_t(X_t^i) + u_t(X_t^i)) \Delta t + \sqrt{2 \sigma_t \Delta t} \, \epsilon_t^i, \quad \epsilon_t^i \sim \mathcal{N}(0, I)$
\EndFor
\State \Return $\{X_1^i\}_{i=1}^N$
\end{algorithmic}
\end{algorithm}

\section{Connections to prior work} \label{sec:prior}

We give an overview of inference-time steering methods in Section~\ref{sec:gmc}, and relate our framework to prior work by discussing alternative approaches to the corrective drifts and $p_t$ invariant correctors in Sections~\ref{sec:ut} and~\ref{sec:correctors}, respectively.

\subsection{Guidance and Monte Carlo approximations} \label{sec:gmc}

\paragraph{Guidance methods.} Sampling from a reward-tilted generative prior can be achieved either by finetuning the model \citep{venkatraman2024amortizing, domingo2025adjoint, potaptchik2026tilt}, or by modifying the generation process at test time. 
Training-free guidance methods \citep{chung2023diffusion, song2023pseudoinverseguided, pokle2024trainingfree, kim2025flowdps} adapt the generative ODE/SDE by adding the reward gradient, typically evaluated at the denoised estimate, but do not sample from the target distribution due to the approximations involved. \citet{nguyen2026stein} consider correcting the distribution of denoised estimates using Stein variational gradient descent (SVGD), but this relies on the prior score evaluated at clean data where it is unreliable.

\paragraph{Monte Carlo methods.} SMC-based methods correct for the bias in guidance through weighted particles and resampling \citep{wu2023practical, dou2024diffusion, singhal2025a, skreta25a, kelvinius25b}.
Alternatively, when flow maps trained for one-step sampling of the final state given an intermediate noisy state are available, a consistent Monte Carlo estimate of the guidance drift itself can be obtained \citep{potaptchik2026meta, holderrieth2026diamond, pan2026maps}.
Markov chain Monte Carlo (MCMC) methods that sample the pullback of the reward-tilted prior onto the base also target the correct distribution, and are asymptotically exact in the large time limit \citep{kalaivanan2025ess, wang2026source}.

\subsection{Parametric approximations of the corrective drift} \label{sec:ut}

\paragraph{Neural networks.} In the related task of sampling unnormalized probability densities, weight degeneracy is mitigated by learning additional transport, where the corrective drift is parameterized by a neural network \citep{arbel2021annealed, vargas2024transport, albergo25a}.
As a specific example, NETS \citep{albergo25a} parameterizes both the drift $u(x,t)$ and the log-partition function $F(t) = \log Z_t = \log Z_0 + \int_0^t \mathbb{E}_{p_s}[g_s] \diff s$ with a network, trained over simulated trajectories to minimize
\begin{equation}
    \mathcal{L}(u, F) = \int_0^1 \mathbb{E}_{p_t} [(S_{p_t}u(t, x) + g_t(x) - \partial_t F(t))^2] \diff t.
\end{equation}
For fixed $t$, this coincides with the IPG-CV objective up to regularization, where $\mu_t$ denotes $\partial_t F(t)$.

\paragraph{Predefined bases.} DriftLite \citep{ren2026driftlite} adapts NETS for reward-tilting by minimizing objective~\eqref{eq:ipsobjective}, without regularization, over drifts of the form
\begin{equation}
    u_t(x) = \phi_1 v_t(x) + \phi_2 \nabla \log q_t(x) + \phi_3 \nabla r(x, t).
\end{equation}
Approximating the expectations in the objective with the weighted empirical distribution yields a closed-form solution by solving a $3 \times 3$ linear system.
However, it requires evaluating the Stein operator $S_{p_t}u_t$ at the particles and involves $\nabla \cdot v_t$, $\Delta \log q_t$ and $\Delta r$. 
These are coarsely approximated with Hutchinson's trace estimator, and as the corrective drift does not drive the reweighting term close to zero, reweighting and resampling are necessary.

\paragraph{Gradient field.} Restricting the corrective drift to a gradient field $u_t(x) = \nabla \psi_t(x)$ for $\psi_t \colon \mathbb{R}^d \rightarrow \mathbb{R}$ allows using an alternate objective
\begin{equation} \label{eq:localAM}
    \mathcal{L}(\psi_t) = \mathbb{E}_{p_t} \bigg[\frac{1}{2} \|\nabla \psi_t(x)\|^2 - \psi_t(x)(g_t(x) - \mathbb{E}_{p_t}[g_t]) \bigg].
\end{equation}
This has been explored for $\psi_t$ parameterized by a neural network \citep{albergo25a} and as a linear combination over predefined bases \citep{ren2026driftlite}.
However, these do not perform as well as directly parameterizing $u_t$.
The objective~\eqref{eq:localAM} also admits a closed-form solution for $\psi_t \in \mathcal{H}_k$, which resembles the kernel Fisher--Rao flow of \citet{maurais24a} for sampling unnormalized probability densities, similar to how IPG relates to Stein transport.

\subsection{Predictor-corrector methods} \label{sec:correctors}

In SMC methods, resample-move \citep{gilks2001following} rejuvenates particles after resampling by applying a $p_t$-invariant MCMC kernel.
The Langevin terms in equation~\eqref{eq:sdetilt} can be seen as an infinitesimal such move applied jointly with the transport rather than as a separate step as in predictor-corrector methods \citep{song2020score}.
Consequently, we can replace the Langevin terms by any continuous time process that leaves $p_t$ invariant, and apply any number of such corrector steps at a fixed time $t$ after resampling in SMC, or directly in the case of IPG.

\paragraph{Interacting correctors.} More generally, we can consider the $N$-particle system on the augmented state space $\bar{x} = (x^1, \ldots , x^N)$ with $p_t^N(\bar{x}) = \prod_{i=1}^N p_t(x^i)$, and use processes that leave $p_t^N$ invariant as correctors.
This includes interacting dynamics such as affine invariant Langevin dynamics \citep{garbuno2020affine} and suitably noise-perturbed SVGD \citep{gallego2018stochastic, nusken2021stein}.
In Appendix~\ref{sec:gmmsvgd}, we evaluate using noisy SVGD in place of the Langevin terms.

\section{Experiments} \label{sec:experiments}

We evaluate our interacting particle guidance (IPG) and its control variate version (IPG-CV), as described in Algorithm~\ref{alg:IPG}, on high-dimensional Gaussian mixture models (GMMs) with known posteriors, and on image inpainting and protein structure inference problems.
In all our experiments, we choose the radial basis function (RBF) kernel, with the bandwidth set using the median heuristic at each time step.
We compare against methods designed to be consistent, such as Feynman--Kac correctors (FKC) \citep{skreta25a} which simulates the weighted SDE in equation~\eqref{eq:sdetilt}, and DriftLite \citep{ren2026driftlite} which adds a corrective drift restricted to predefined bases.

\subsection{Gaussian mixture models}

We consider a 256-dimensional GMM as the prior $q_1(x)$, and a linear-Gaussian observation $y=Ax + \sigma_\text{obs} \epsilon$, where $y \in \mathbb{R}^{128}$.
To sample from the posterior $p_1(x) = q_1(x|y)$, we set the log-likelihood as the reward $R(x)$, and propagate particles along the interpolation $p_t(x) \propto q_t(x) \exp{(tR(x))}$ from $t=0$ to $t=1$.
The intermediate marginal densities $q_t(x)$ correspond to those of a noising Ornstein--Uhlenbeck (OU) process, whose time-reversed SDE is known.
We use the associated probability flow ODE as the generative process instead of training a flow-based model, and compare samples against the exact posterior, which is available in closed form.

\begin{figure}[t]
\begin{center}
\includegraphics[width=0.7\linewidth]{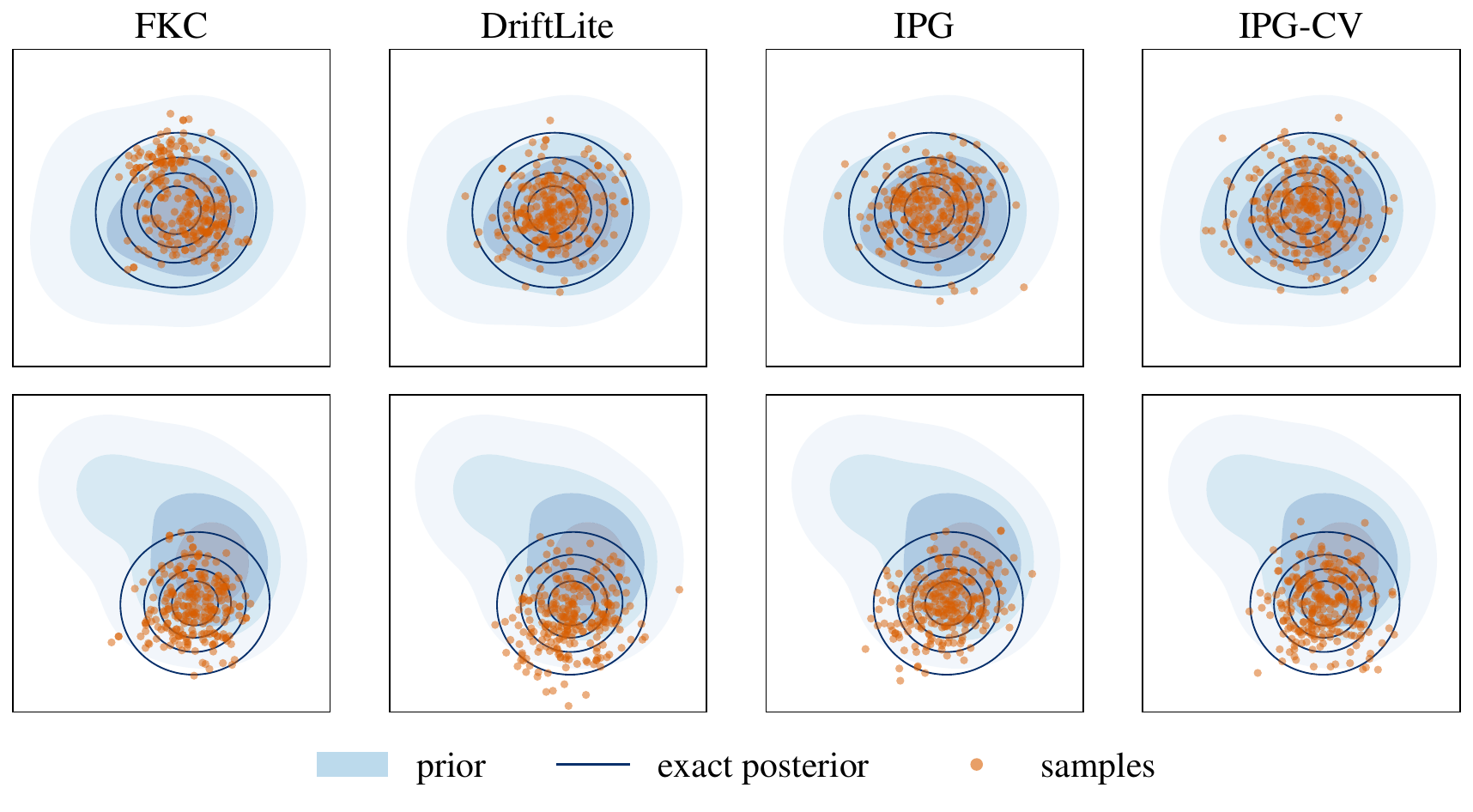}
\end{center}
\caption{Samples on random 2-D slices of the 256-dimensional posterior.}
\label{fig:gmm}
\end{figure}

We evaluate FKC and DriftLite under different resampling policies, and compare them against both versions of interacting particle guidance, which do not require resampling.
We run the methods with $N=256$ particles and report the effective sample size (ESS), mean error, maximum mean discrepancy (MMD), and sliced 2-Wasserstein distance (SWD), over $5$ randomly generated priors and observations in Table~\ref{tab:gmm}.
Figure~\ref{fig:gmm} compares samples from FKC and DriftLite, both using systematic resampling at every step, with those from IPG and IPG-CV.
Further details on the hyperparameters and metrics can be found in Appendix~\ref{sec:gmmappendix}. 
Additional experiments replacing the Langevin terms with noisy SVGD \citep{gallego2018stochastic, nusken2021stein}, and using a neural corrective drift, are in Appendices~\ref{sec:gmmsvgd} and~\ref{sec:gmmneural}, respectively.

\begin{table}[t]
\centering
\small
\caption{Mean and standard deviation of the metrics on the GMM experiment. We use systematic resampling and ``Adaptive'' refers to resampling whenever ESS/$N$ falls below $0.5$.}
\label{tab:gmm}
\begin{tabular}{@{}llcccc@{}}
\toprule
Method & Resampling & ESS/$N$ $\uparrow$ & Mean error $\downarrow$ & MMD $\downarrow$ & SWD $\downarrow$ \\
\midrule
\multirow{3}{*}{FKC}
 & None       & 0.006 $\pm$ 0.002 & 7.593 $\pm$ 1.337 & 0.876 $\pm$ 0.165 & 0.682 $\pm$ 0.112 \\
 & Adaptive   & 0.771 $\pm$ 0.203 & 4.485 $\pm$ 0.872 & 0.184 $\pm$ 0.042 & 0.311 $\pm$ 0.055 \\
 & Every step & 1.000 $\pm$ 0.000 & 3.685 $\pm$ 0.741 & 0.148 $\pm$ 0.021 & 0.260 $\pm$ 0.043 \\
\midrule
\multirow{3}{*}{DriftLite}
 & None       & 0.011 $\pm$ 0.004 & 5.691 $\pm$ 1.194 & 0.619 $\pm$ 0.139 & 0.509 $\pm$ 0.092 \\
 & Adaptive   & 0.864 $\pm$ 0.169 & 2.178 $\pm$ 0.357 & 0.097 $\pm$ 0.007 & 0.168 $\pm$ 0.021 \\
 & Every step & 1.000 $\pm$ 0.000 & 2.483 $\pm$ 0.243 & 0.085 $\pm$ 0.004 & 0.182 $\pm$ 0.012 \\
\midrule
IPG           & None & -- & \textbf{0.841 $\pm$ 0.051} & \textbf{0.012 $\pm$ 0.002} & \textbf{0.093 $\pm$ 0.002} \\
IPG-CV        & None & -- & 0.842 $\pm$ 0.031 & 0.013 $\pm$ 0.002 & \textbf{0.093 $\pm$ 0.002} \\
\bottomrule
\end{tabular}
\end{table}

Without resampling, the baselines suffer from severe weight degeneracy, with one to three effective particles out of $256$.
While resampling at each step trivially yields ESS/$N = 1$, it does not imply accurate samples as shown by the other metrics. 
With a small regularization constant, the weight updates in IPG and IPG-CV are negligible, and they outperform the baselines across every metric.

\subsection{Image inpainting}

We now evaluate the methods on inpainting high-resolution ($256 \times 256$) images, using a flow-based model \citep{liu2022flow} trained on AFHQ Cat \citep{choi2020stargan} as the generative prior $q_1(x)$.
Given a noisy observation $y = Ax + \sigma_\text{obs} \epsilon$, we aim to sample the posterior $p_1(x) = q_1(x|y)$, which corresponds to a reward-tilted prior with the log-likelihood as the reward $R(x)$.
Here, $A$ denotes an operator that either applies a box mask at the center of the image or masks the entire right half of the image.
We propagate particles along the interpolation $p_t(x) \propto q_t(x) \exp(t R(\hat{x}_t))$ from $t=0$ to $t=1$, where $\hat{x}_t = x_t + (1-t) v_t(x_t)$ is the denoised estimate.

All methods use $N=16$ particles, limited by GPU memory.
Unlike in the GMM case, evaluating $S_{p_t} u_t(x)$ exactly for DriftLite is prohibitively expensive, as it requires $\nabla \cdot v_t(x), \Delta \log q_t(x)$ and $\Delta r(x, t)$, which we approximate using Hutchinson's trace estimator.
Following \citet{skreta25a}, we apply systematic resampling at each time step for both FKC and DriftLite.

\begin{figure}[t]
\begin{center}
\includegraphics[width=0.8\linewidth]{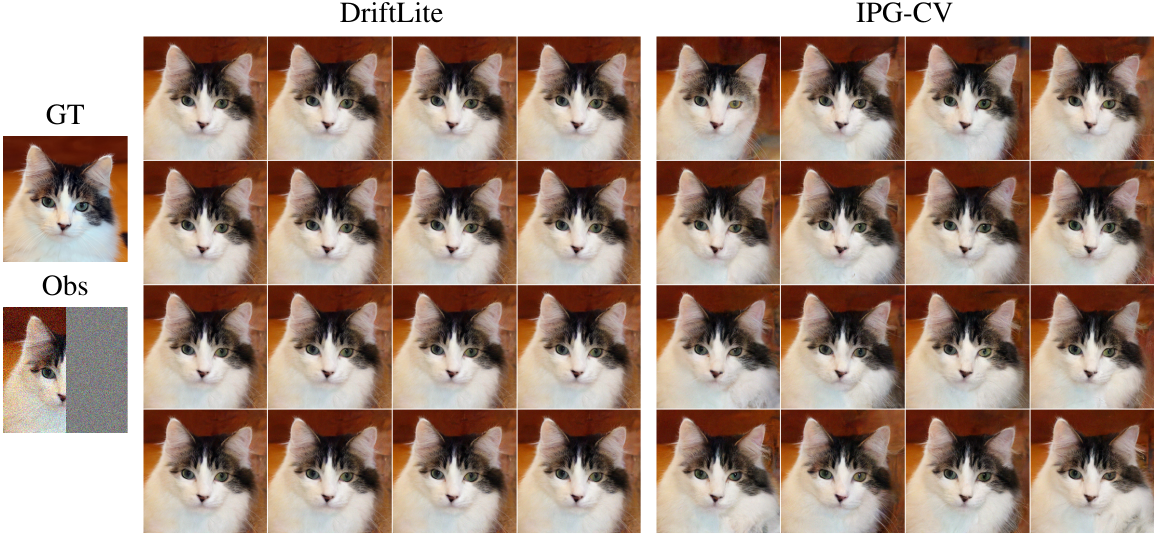}
\end{center}
\caption{DriftLite suffers from particle collapse, producing indistinguishable samples, while IPG-CV produces samples with perceptible differences.}
\label{fig:outpaint}
\end{figure}

We report the commonly used performance metrics, peak signal-to-noise ratio (PSNR), structural similarity index measure (SSIM) and learned perceptual image patch similarity (LPIPS).
Additionally, we measure the diversity among the samples generated per observation using the trace of the sample covariance in the raw pixel space and the mean pairwise LPIPS.
The average values of these metrics, computed over $50$ test images, are given in Table~\ref{tab:img}.
Samples from DriftLite are compared against those from IPG-CV in Figure~\ref{fig:outpaint}.
Further details on the metrics and additional visualizations of the generated samples are provided in Appendix~\ref{sec:imageappendix}.

\begin{table}[t]
\centering
\small
\caption{Results on the image inpainting tasks with $N=16$ particles. ``Avg.'' refers to the mean of the metrics computed between each of the $N$ samples and the ground truth, and ``Best'' takes the best-of-$N$ sample metrics before averaging across the test images.}
\label{tab:img}
\begin{tabular}{@{}llccccccrc@{}}
\toprule
 &  & \multicolumn{2}{c}{PSNR $\uparrow$} & \multicolumn{2}{c}{SSIM $\uparrow$} & \multicolumn{2}{c}{LPIPS $\downarrow$} & \multicolumn{2}{c}{Diversity $\uparrow$} \\
\cmidrule(lr){3-4} \cmidrule(lr){5-6} \cmidrule(lr){7-8} \cmidrule(lr){9-10}
Mask & Method & Avg. & Best & Avg. & Best & Avg. & Best & Cov.\ trace & LPIPS div. \\
\midrule
\multirow{4}{*}{Half}
 & FKC       & \textbf{15.99} & 15.99 & \textbf{0.4534} & 0.4537 & \textbf{0.3352} & 0.3341 & 24.61 & 0.0041 \\
 & DriftLite & 15.51 & 15.52 & 0.4392 & 0.4401 & 0.3466 & 0.3443 & 70.56 & 0.0125 \\
 & IPG       & 15.74 & \textbf{17.98} & 0.4427 & \textbf{0.4894} & 0.3455 & \textbf{0.3058} & \textbf{7990.21} & 0.2307 \\
 & IPG-CV    & 15.68 & 17.79 & 0.4418 & 0.4889 & 0.3462 & 0.3060 & 7742.43 & \textbf{0.2311} \\
\midrule
\multirow{4}{*}{Box}
 & FKC       & \textbf{24.53} & 24.53 & 0.6737 & 0.6739 & \textbf{0.1866} & 0.1859 & 14.02 & 0.0020 \\
 & DriftLite & 24.45 & 24.45 & \textbf{0.6742} & \textbf{0.6750} & 0.1884 & 0.1863 & 46.97 & 0.0065 \\
 & IPG       & 24.29 & 25.22 & 0.6624 & 0.6708 & 0.1981 & 0.1830 & 1005.08 & 0.0957 \\
 & IPG-CV    & 24.26 & \textbf{25.28} & 0.6622 & 0.6711 & 0.1994 & \textbf{0.1819} & \textbf{1028.86} & \textbf{0.0964} \\
\bottomrule
\end{tabular}
\end{table}

The diversity metrics indicate that FKC and DriftLite suffer from severe particle collapse, producing near identical samples, whereas IPG and IPG-CV yield perceptually diverse samples.
While the baselines achieve marginally higher average metrics, we note that the mean PSNR, SSIM and LPIPS computed between the generated samples and ground truth image are imperfect measures of posterior sample quality.
Samples from a diverse posterior might yield worse average scores than collapsed samples that happen to be close to the ground truth.
We therefore also report best-of-$N$ metrics, which show that IPG and IPG-CV produce samples that fit the ground truth better as well.

We note that both IPG and IPG-CV incur negligible computational overhead ($122.9$s per test image) compared to FKC ($122.1$s per test image).
This is because the cost of the $N\times N$ linear solve is low for the typical number of particles used when sampling reward-tilted generative priors, and the matrix entries only require the kernel, its gradients, and terms already used in the particle updates.
In contrast, DriftLite requires evaluating Hutchinson estimates of $S_{p_t}u_t(x)$ at the particle positions and is much slower ($444.3$s per test image). 

\subsection{Protein structures from pairwise distance measurements}

We compare the methods on inferring the backbone structure of $M$-residue proteins from inter-residue distances, using the flow-based model Proteina \citep{geffner2025proteina} as the generative prior $q_1(x)$.
The backbone structure $x = (x_1, \ldots, x_M) \in \mathbb{R}^{M \times 3}$ is given by the position of $\alpha$-carbon in each residue, and the observation $y = \mathcal{A}(x) + \sigma_{\text{obs}} \epsilon \in \mathbb{R}^K$ consists of $K$ noisy pairwise distances, $[\mathcal{A}(x)]_k = \|x_{i_k} - x_{j_k}\|_2$ for distinct pairs $(i_k, j_k)$ of residues drawn uniformly at random, with $K$ as $3\%$ of all possible pairs.
As in the image inpainting task, we sample the posterior $p_1(x) = q_1(x|y)$, which corresponds to setting the log-likelihood as the reward $R(x)$, by propagating particles along the interpolation $p_t(x) \propto q_t(x) \exp(t R(\hat{x}_t))$.
Here, $\hat{x}_t = x_t + (1-t) v_t(x_t)$ is the denoised estimate.

As the observation operator depends only on pairwise distances, which are invariant to rotations, and Proteina is trained with random rotations as data augmentation, the orientation of the structures is uniform under the posterior. 
This is a deliberate choice which lets us test whether the methods preserve this diversity.
We also note that Proteina produces designable structures at a rate below $20\%$, unless heuristic noise reduction is applied during generation \citep{geffner2025proteina}.
As we focus on evaluating inference-time steering, we use the unmodified prior for all methods and accept the low designability as a byproduct.

All methods use $N=16$ particles.
As in the image inpainting task, the divergences required for DriftLite are approximated with Hutchinson's trace estimator, and systematic resampling is applied at each time step for both FKC and DriftLite.
We measure accuracy using the root mean square deviation (RMSD) between the generated structures $\{x^i\}_{i=1}^N$ and the ground truth structure after rotational alignment ($\text{RMSD}_\text{gt}$), and the RMSD between the pairwise distances in the generated samples $\{\mathcal{A}(x^i)\}_{i=1}^N$ and the observation $y$ ($\text{RMSD}_\text{obs}$).
To measure the diversity, for each pair of generated structures, we compute the angle of rotation needed to align them in degrees (Rot. div.) and the RMSD between them after this alignment (RMSD div.).
The expected Rot. div. is $126.5^\circ$, which is the average angle of rotation between two independent random orientations of a structure.
Following standard practice, a structure is considered designable when its self-consistency RMSD (see Appendix~\ref{sec:proteinappendix}) is below $2$ \AA.
We report the fraction of test cases for which a method generates at least one designable structure.
The average metrics computed over $15$ protein structures are given in Table~\ref{tab:protein}.
Further details on the test proteins and evaluation metrics, along with visualizations of the generated samples, are provided in Appendix~\ref{sec:proteinappendix}.

\begin{table}[t]
\centering
\small
\caption{Results on the protein structure inference task with $N=16$ particles. ``Avg.'' refers to the mean of the metrics computed between each of the $N$ samples and the ground truth, and ``Best'' takes the best-of-$N$ sample metrics for each test protein before averaging.}
\label{tab:protein}
\begin{tabular}{@{}lccccrrc@{}}
\toprule
 & \multicolumn{2}{c}{$\text{RMSD}_\text{obs}$ (\AA) $\downarrow$} & \multicolumn{2}{c}{$\text{RMSD}_\text{gt}$ (\AA) $\downarrow$} & \multicolumn{2}{c}{Diversity $\uparrow$} & \\
\cmidrule(lr){2-3} \cmidrule(lr){4-5} \cmidrule(lr){6-7}
Method & Avg. & Best & Avg. & Best & Rot.\ div.\ ($^\circ$) & RMSD div.\ (\AA) & Any designable $\uparrow$ \\
\midrule
FKC       & 1.225 & 1.200 & 8.119 & 8.060 & 0.6 & 0.677 & 7/15 \\
DriftLite & \textbf{1.188} & 1.163 & \textbf{6.907} & 6.840 & 0.7 & 0.756 & 8/15 \\
IPG       & 1.308 & 1.053 & 7.881 & 1.298 & 122.1 & \textbf{7.981} & 13/15 \\
IPG-CV    & 1.290 & \textbf{1.048} & 7.543 & \textbf{1.232} & \textbf{126.9} & 7.800 & \textbf{15/15} \\
\bottomrule
\end{tabular}
\end{table}

\begin{figure}[t]
\begin{center}
\includegraphics[width=0.9\linewidth]{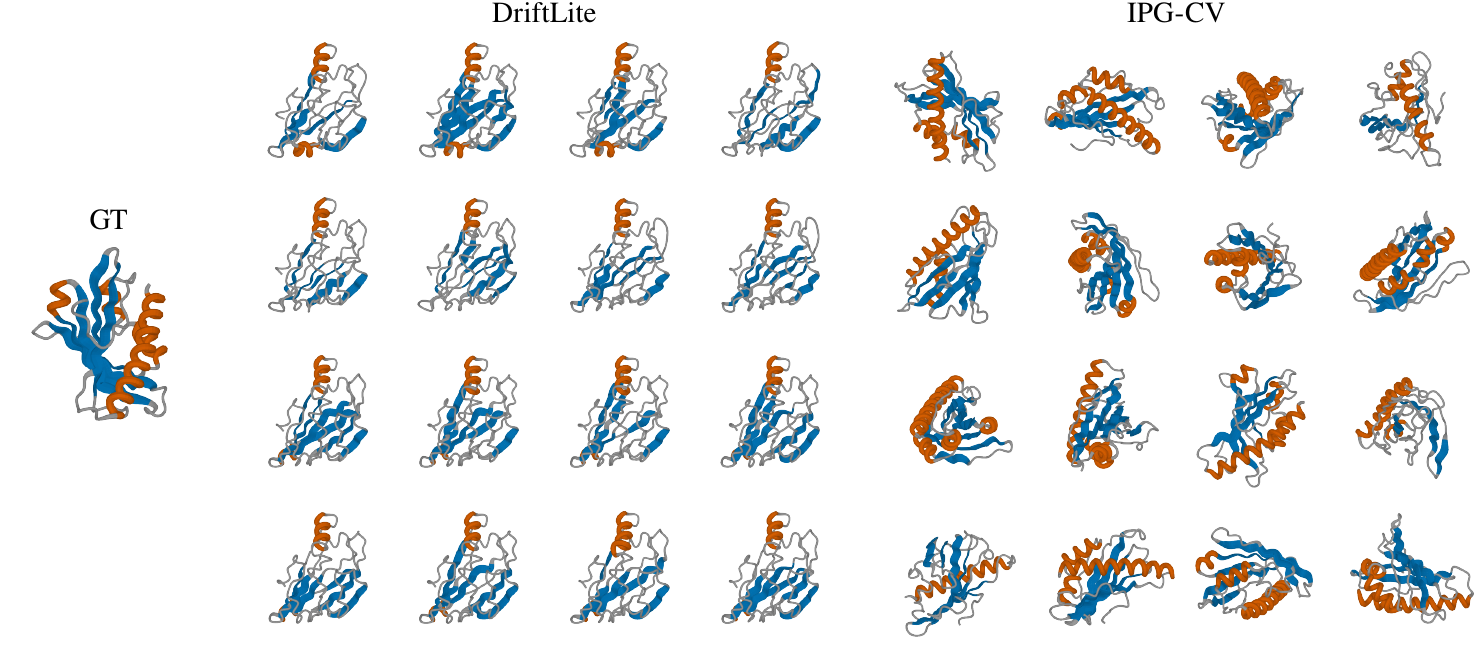}
\end{center}
\caption{Samples produced on the backbone structure inference task for protein \texttt{30JI}. They are intentionally not aligned with the ground truth to highlight the particle collapse in DriftLite. Samples from IPG-CV contain structures close to the ground truth, and have random orientations as expected.}
\label{fig:protein}
\end{figure}

Both the diversity metrics in Table~\ref{tab:protein} and the samples visualized in Figure~\ref{fig:protein} show that FKC and DriftLite suffer from severe particle collapse, producing structures with near identical shapes and orientations.
In contrast, both IPG and IPG-CV produce samples with diverse orientations, with Rot. div. close to the expected value of $126.5^\circ$.
The average $\text{RMSD}_\text{gt}$ is high for all the methods, which is consistent with the low designability of the Proteina prior.
However, the best-of-$N$ $\text{RMSD}_\text{gt}$ shows that both IPG and IPG-CV generate structures close to the ground truth, and IPG-CV produces at least one designable structure in every test case.
The baselines, on the other hand, collapse around structures that are often undesignable, resulting in a much higher best-of-$N$ $\text{RMSD}_\text{gt}$.
In terms of runtime, IPG and IPG-CV ($126.0$s per test structure) are nearly as fast as FKC ($125.7$s per test structure), while DriftLite is much slower ($460.6$s per test structure). 

\section{Conclusion}

We introduce interacting particle guidance (IPG), a training-free method for sampling reward-tilted generative priors, in which an interacting corrective drift is derived to cancel the reweighting term in the Feynman--Kac PDE.
Solving for the drift in an RKHS yields a closed-form solution with negligible computational overhead compared to existing SMC methods.
We evaluate IPG on high-dimensional Gaussian mixtures, image inpainting and protein structure inference problems, and show significant improvements over prior work.
Specifically, we see improvements over SMC-based formulations, despite the fact that IPG and SMC/FKC are derived from the same Feynman--Kac PDE. We conjecture that the continuous ``nudging'' of particles in combination with Langevin-style guidance allows IPG to track the evolution of $p_t$ closely despite finite-sample errors in the forces. This is in contrast with a pure SMC-based method where any lag in the sampling dynamics with respect to $p_t$ accumulates in the importance weights until they degenerate or trigger resampling.

Despite the compelling empirical performance, like any sampling algorithm, the method has practical limitations.
By replacing reweighting and resampling with transport, the method avoids particle collapse, but may struggle to move particles across large energy barriers in a multi-modal distribution when the intermediate reward gradients $\|\nabla r(x,t)\|$ are much weaker than the prior score $\|\nabla \log q_t(x)\|$.
This could be remedied either by tuning the interpolating path $p_t$, or by increasing the regularization constant $\lambda$, which reintroduces particle weights, and thereby resampling, while retaining the variance-reducing corrective transport.
Here, we demonstrated the method using a simple RBF kernel, which performs well but is agnostic to the data.
A promising direction for future work is to use learned kernels \citep{galashov2025deep} or to operate in a pretrained latent space, either of which could improve the performance further.

\subsection*{AI use statement}

In this work, we used coding agents to implement the methods used in our experiments, and verified the outputs for correctness.
Additionally, we used generative AI tools to polish the writing, with minor revisions to grammar and phrasing.
We did not use generative AI tools for any other tasks that require disclosure.
We take full responsibility for the final content of this work, including all text and artifacts produced with the aid of generative AI.

\subsection*{Reproducibility statement}

We clearly describe our algorithm in Section~\ref{sec:method}.
Details of our experiments, including the models, datasets, and hyperparameters used, are described in Section~\ref{sec:experiments} and the appendix.
Our code is not included with this submission, but will be made publicly available upon publication.

\subsubsection*{Acknowledgments}
This work was financially supported by the Wallenberg AI, Autonomous Systems and Software Program (WASP) funded by the Knut and Alice Wallenberg Foundation,
the Swedish Research Council (project no: 2024-05011),
and the Excellence Center at Linköping--Lund in Information Technology (ELLIIT). Computational resources were provided on the Berzelius system funded by the Knut and Alice Wallenberg foundation and operated by NAISS.

\bibliography{iclr2027_conference}
\bibliographystyle{iclr2027_conference}

\clearpage
\appendix
\section{Reward-tilted Feynman--Kac PDE} \label{sec:rtfkp}
For the interpolation $p_t(x) \propto q_t(x) \exp(r(x, t))$ defined in Section~\ref{sec:rewardtilting}, with the normalizing constant $Z_t = \int_{\mathbb{R}^d} q_t(x) \exp(r(x, t)) \diff x$, we have
\begin{equation}
    \log p_t(x) = \log q_t(x) + r(x, t) - \log Z_t.
\end{equation}
Differentiating with respect to time gives
\begin{equation}
\begin{split}
    \frac{\partial_t p_t(x)}{p_t(x)} &= \frac{\partial_t q_t(x)}{q_t(x)} + \partial_t r(x, t) - \partial_t \log Z_t \\
    &= - \nabla \cdot v_t(x) - \langle v_t(x), \nabla \log q_t(x) \rangle + \partial_t r(x, t) - \partial_t \log Z_t \\
    &= - \nabla \cdot v_t(x) - \langle v_t(x), \nabla \log p_t(x) - \nabla r(x, t) \rangle + \partial_t r(x, t) - \partial_t \log Z_t,
\end{split}
\end{equation}
where we use the continuity equation for $q_t$ and substitute $\nabla \log q_t = \nabla \log p_t - \nabla r(x, t)$. This can be rewritten as
\begin{equation}
    \partial_t p_t(x) = -\nabla \cdot (v_t(x) p_t(x)) + p_t(x)(\partial_t r(x, t) + \langle v_t(x), \nabla r(x, t) \rangle - \partial_t\log Z_t).
\end{equation}
Since $\int_{\mathbb{R}^d} p_t(x) \diff x = 1$ for all $t$, differentiating both sides with respect to $t$ gives $\int_{\mathbb{R}^d} \partial_t p_t(x) \diff x = 0$, which yields
\begin{equation}
    \partial_t\log Z_t = \mathbb{E}_{p_t}[\partial_t r(x, t) + \langle v_t(x), \nabla r(x, t) \rangle].
\end{equation}
Substituting the expression for $\partial_t\log Z_t$ back, and defining $g_t(x) = \partial_t r(x, t) + \langle v_t(x), \nabla r(x, t) \rangle$, we obtain
\begin{equation}
    \partial_t p_t(x) = - \nabla \cdot (v_t(x) p_t(x)) + p_t(x) (g_t(x) - \mathbb{E}_{p_t}[g_t]).
\end{equation}

\section{Derivation of IPG and IPG-CV drifts} \label{sec:ipgcv}

We obtain the minimizer of objective~\eqref{eq:ipsobjective} for a general regularization constant $\lambda > 0$ with weighted particles $\{(X_t^i, w_t^i)\}_{i=1}^N$, adapting the proof from \citet[Appendix A.1]{nusken2024stein}.
In both IPG and IPG-CV, we choose a small $\lambda$ such that the weight updates can be ignored, and setting $w_t^i = 1/N$ recovers equations~\eqref{eq:rkhsdrift} and~\eqref{eq:rkhscdrift}.

Let $W = \mathrm{diag}(w_t) \in \mathbb{R}^{N \times N}$, and $\langle x, y \rangle_{w_t} = \sum_{i=1}^N w_t^i x_i y_i$ denote the weighted inner product on $\mathbb{R}^N$. 
We define the sample Stein operator $S_{p_t,N}\colon \mathcal{H}_k^d \rightarrow \mathbb{R}^N$ by $(S_{p_t,N}u_t)_i = S_{p_t}u_t(X_t^i)$, the weighted centering operator $\Pi_w = I_N - \mathbf{1}w_t^\top$, and the vector $\mathrm{g}_t \in \mathbb{R}^N$ by $[\mathrm{g}_t]_i = g_t(X_t^i) - \sum_{j=1}^N w_t^j g_t(X_t^j)$.
Approximating $\mathbb{E}_{p_t}[\cdot]$ in objective~\eqref{eq:ipsobjective} with the weighted particles, we get the empirical version
\begin{equation}
    \hat{\mathcal{L}}(u_t) = \| S_{p_t, N} u_t + \mathrm{g}_t \|_{w_t}^2 + \lambda \| u_t \|_{\mathcal{H}_k^d}^2.
\end{equation}
For IPG-CV, using the optimal $\mu_t = \sum_{i=1}^N w_t^i (S_{p_t} u_t(X_t^i) + g_t(X_t^i))$ yields
\begin{equation}
    \hat{\mathcal{L}}(u_t) = \| \Pi_w S_{p_t, N} u_t + \mathrm{g}_t \|_{w_t}^2 + \lambda \| u_t \|_{\mathcal{H}_k^d}^2.
\end{equation}
Therefore, both the objectives are of the form
\begin{equation} \label{eq:localpinnrkhs}
    \hat{\mathcal{L}}(u_t) = \| P S_{p_t, N} u_t + \mathrm{g}_t \|_{w_t}^2 + \lambda \| u_t \|_{\mathcal{H}_k^d}^2,
\end{equation}
with $P = I_N$ for IPG and $P = \Pi_w$ for IPG-CV.

Following the Tikhonov regression approach of \citet[Appendix A.1.2]{nusken2024stein}, the minimizer of equation~\eqref{eq:localpinnrkhs} is
\begin{equation} \label{eq:rkhscderivation}
    u_t = (P S_{p_t, N})^*(\lambda I_N + (P S_{p_t, N})(P S_{p_t, N})^*)^{-1}(-\mathrm{g}_t),
\end{equation}
where the adjoint is taken with respect to $\langle \cdot, \cdot \rangle_{w_t}$ and the inner product of $\mathcal{H}_k^d$.
For any $x, y \in \mathbb{R}^N$,
\begin{equation}
    \langle x, \Pi_w y \rangle_{w_t} = \sum_{i=1}^N w_t^i x_i y_i - \big(\sum_{i=1}^N w_t^i x_i\big) \big( \sum_{j=1}^N w_t^j y_j \big) = \langle \Pi_w x, y \rangle_{w_t},
\end{equation}
from which it follows that $\Pi_w$ is self-adjoint, and in both cases $P^* = P$ and $(P S_{p_t, N})^* = S_{p_t, N}^* P$.
The adjoint $S_{p_t, N}^*\colon \mathbb{R}^N \rightarrow \mathcal{H}_k^d$, satisfying $\langle c, S_{p_t, N} v \rangle_{w_t} = \langle S_{p_t, N}^* c, v \rangle_{\mathcal{H}_k^d}$ for all $v \in \mathcal{H}_k^d$ and $c \in \mathbb{R}^N$, is given by \citet[Remark 11]{nusken2024stein}
\begin{equation}
    S_{p_t, N}^*c = \sum_{j=1}^N w_t^j c_j [k(\cdot, X_t^j) \nabla \log p_t(X_t^j) + \nabla_{X_t^j} k(\cdot, X_t^j)].
\end{equation}
By direct calculation, $(S_{p_t, N}S_{p_t, N}^* c)_i = \sum_{j=1}^N \xi_{ij} w_t^j c_j$, i.e., $S_{p_t, N} S_{p_t, N}^* = \xi W$, and
\begin{equation}
    (P S_{p_t, N})(P S_{p_t, N})^* = P S_{p_t, N} S_{p_t, N}^* P =  P \xi W P.
\end{equation}
Using these results, equation~\eqref{eq:rkhscderivation} can be rewritten as
\begin{equation} \label{eq:weightedsolution}
    u_t = S_{p_t, N}^* P \phi, \quad (P \xi W P + \lambda I_N) \phi = -\mathrm{g}_t.
\end{equation}
For IPG, this gives $u_t(x) = \sum_{j} w_t^j \phi_j [k(x, X_t^j) \nabla \log p_t(X_t^j) + \nabla_{X_t^j} k(x, X_t^j)]$ with $(\xi W + \lambda I_N)\phi = -\mathrm{g}_t$, and for IPG-CV, the same $u_t$ with $\phi$ replaced by $\Pi_w \phi$, and $(\Pi_w \xi W \Pi_w + \lambda I_N)\phi = -\mathrm{g}_t$.
Setting $w_t^i = 1/N$, such that $W = I_N/N$ and $\Pi_w = \Pi$, recovers our equations~\eqref{eq:rkhsdrift} and~\eqref{eq:rkhscdrift}.

On expanding the Stein operators, the entries in the Gram matrix are given by,
\begin{equation}
\begin{split}
    [\xi]_{ij} = \langle \nabla \log p_t(X_t^i), \nabla_{X_t^j}k(X_t^i, X_t^j) \rangle + \langle \nabla \log p_t(X_t^j), \nabla_{X_t^i}k(X_t^i, X_t^j) \rangle \\ + k(X_t^i, X_t^j) \langle \nabla \log p_t(X_t^i), \nabla \log p_t(X_t^j) \rangle + \nabla_{X_t^i} \cdot \nabla_{X_t^j}k(X_t^i, X_t^j),
\end{split}
\end{equation}
and are cheap to compute as they involve the kernel, its gradients, and terms already used in the particle updates.

\section{Adding diffusion-reweighting}

The methods discussed so far are motivated by the different degrees of freedom in the Feynman--Kac PDE~\eqref{eq:partialpt}, such as adding drift-diffusion and drift-reweighting terms, and so it is natural to consider adding diffusion-reweighting terms. 

For a symmetric positive semi-definite diffusion tensor $D_t(x) \in \mathbb{R}^{d \times d}$, let us add and subtract $\nabla \cdot \nabla \cdot (D_t(x) p_t(x))$ in equation~\eqref{eq:partialpt}, and write $- \nabla \cdot \nabla \cdot (D_t(x) p_t(x)) = - p_t(x)(\mathcal{A}_{p_t}D_t(x))$ where
\begin{equation}
\begin{split}
    \mathcal{A}_{p_t}D_t(x) &= \nabla \cdot \nabla \cdot D_t(x) + 2 \langle \nabla \cdot D_t(x), \nabla \log p_t(x) \rangle + \langle D_t(x), \nabla^2 \log p_t(x) \rangle_F \\
    &\quad + \nabla \log p_t(x)^\top D_t(x) \nabla \log p_t(x),
\end{split}
\end{equation}
where $(\nabla \cdot D_t(x))_i = \sum_{j=1}^d \partial_{x_j}D_t(x)_{ij}$, $\nabla \cdot \nabla \cdot D_t(x) = \sum_{i, j = 1}^d \partial_{x_i} \partial_{x_j} D_t(x)_{ij}$ and $\langle \cdot, \cdot \rangle_F$ denotes the Frobenius inner product. The modified PDE for $p_t$ is given by
\begin{equation}
    \partial_t p_t(x) = - \nabla \cdot (v_t(x) p_t(x)) + \nabla \cdot \nabla \cdot (D_t(x) p_t(x)) + p_t(x) (g_t(x) - \mathcal{A}_{p_t}D_t(x) - \mathbb{E}_{p_t}[g_t]).
\end{equation}
Following our earlier procedures, $p_t$ can be approximated with a weighted stochastic particle system
\begin{equation}
    \diff X_t^i = v_t(X_t^i) \diff t + \sqrt{2 D_t(X_t^i)} \diff W_t, \quad \diff \omega_t^i = (g_t(X_t^i) - \mathcal{A}_{p_t}D_t(X_t^i)) \diff t.
\end{equation}

As before, $D_t$ can be estimated to minimize the variance of the particle weights. However, since $\nabla \cdot \nabla \cdot (D_t(x) p_t(x)) = -\nabla \cdot (u_t(x) p_t(x))$ for $u_t(x) = - D_t(x) \nabla \log p_t(x) - \nabla \cdot D_t(x)$, diffusion-reweighting can be treated as a specific instance of drift-reweighting, and may not offer any additional capabilities.

\section{Additional details on the GMM experiment} \label{sec:gmmappendix}

\paragraph{Prior.} We consider a 256-dimensional GMM with 10 equally weighted components as the prior $q_1(x)$.
The component means are drawn i.i.d. from $U[-1.3, 1.3]^{256}$, and are then centered so that the mixture mean is zero.
The isotropic per-component variance is chosen such that the total variance of the GMM is $1$.
The intermediate densities $q_t$ correspond to the marginals of the noising OU process at time $s=1-t$,
\begin{equation}
    \diff Z_s = -a Z_s \diff s + b \diff W_s, \quad Z_0 \sim q_1,
\end{equation}
where $a = 3$ and $b = \sqrt{6}$.
The generative process is given by the probability flow ODE associated with the time-reversal of the noising SDE, $\diff X_t = v_t(X_t) \diff t$, initialized as $X_0 \sim q_0$, with
\begin{equation}
    v_t(x) = a x + \frac{b^2}{2} \nabla \log q_t(x).
\end{equation}
In this case, $q_t(x)$ is available in closed form for $t \in [0,1]$.

\paragraph{Reward.} We consider a linear-Gaussian observation $y=Ax + \sigma_\text{obs} \epsilon \in \mathbb{R}^{128}$ with $\sigma^2_{\text{obs}} = 0.1$, for which the likelihood function is $\mathcal{N}(y; Ax, \sigma^2_\text{obs} I)$.
The entries in observation matrix $A \in \mathbb{R}^{128 \times 256}$ are drawn i.i.d. from $\mathcal{N}(0,1)$, and then the rows are unit normalized.
The posterior GMM $p_1(x)$ corresponds to a reward-tilted prior where the reward $R(x)$ is the log-likelihood, and is known analytically.

All methods propagate particles along time $t \in [0,1]$, uniformly discretized into $500$ steps, with a constant Langevin noise scale $\sigma_t = 3$.
For both IPG and IPG-CV, we use $\lambda = 10^{-3}$ and RBF kernel with adaptive bandwidth chosen at each time step as the median of pairwise distances
\begin{equation}
    \sigma_k^2 = \text{median}^2/ \log N.
\end{equation}

\paragraph{Metrics.} The effective sample size (ESS) is computed using the self-normalized particle weights at final time $t=1$.
The other metrics are reported over $256$ unweighted particles obtained after systematic resampling, which has no effect if the particles are equally weighted at $t=1$.
The mean error is the $L_2$ norm of the difference between the empirical sample mean and the exact posterior mean.
$\text{MMD}^2$ is computed using the unbiased estimator, against $M=256$ samples drawn from the exact posterior.
We use an RBF kernel with the bandwidth
\begin{equation}
    \sigma_\text{MMD}^2 = \text{median}^2/(2 \log M),
\end{equation}
where the median is over pairwise distances of the posterior samples.
The sliced 2-Wasserstein distance (SWD) is also reported against the posterior samples with $10^3$ random projections.

\subsection{Replacing overdamped Langevin with noisy SVGD} \label{sec:gmmsvgd}

The equations~\eqref{eq:rkhsdrift} and \eqref{eq:rkhscdrift} resemble the SVGD drift targeting $p_t$ at each time $t$, up to coefficients $\{\phi_j\}_{j=1}^N$.
Therefore, as mentioned in Section~\ref{sec:correctors}, we consider replacing the overdamped Langevin terms in the particle updates with a suitably noise perturbed version of SVGD \citep{gallego2018stochastic, nusken2021stein}.
The resulting modified particle update is given by
\begin{equation}
\begin{split}
    \diff X_t^i = \bigg(v_t(X_t^i) + u_t(X_t^i) + \frac{1}{N} \sum_{j=1}^N [k(X_t^i, X_t^j) \nabla \log p_t(X_t^j) + \nabla_{X_t^j} k(X_t^i, X_t^j)] \bigg) \diff t \\
    + \sum_{j=1}^N \bigg(\sqrt{\frac{2}{N} D_t}\bigg)_{ij} \diff W_t^j,
\end{split}
\end{equation}
where $D_t \in \mathbb{R}^{dN \times dN}$ is a block-structured matrix with the $(i,j)$-th block as $k(X_t^i, X_t^j) I_d$, $(\sqrt{D_t})_{ij}$ denotes the $(i,j)$-th block of the matrix square root, and $W_t^j \in \mathbb{R}^d$ are independent standard Brownian motions.
We use an RBF kernel with unit bandwidth and compare methods with different corrective drifts.
The performance metrics computed over $5$ randomly generated priors and observations are shown in Table~\ref{tab:gmmsvgd}.

\begin{table}[t]
\centering
\small
\caption{Mean and standard deviation of the metrics when overdamped Langevin is replaced with noisy SVGD on the GMM experiment.}
\label{tab:gmmsvgd}
\begin{tabular}{@{}llcccc@{}}
\toprule
Method & Resampling & ESS/$N$ $\uparrow$ & Mean error $\downarrow$ & MMD $\downarrow$ & SWD $\downarrow$ \\
\midrule
\multirow{2}{*}{FKC}
 & None       & 0.006 $\pm$ 0.002 & 7.593 $\pm$ 1.337 & 0.876 $\pm$ 0.165 & 0.682 $\pm$ 0.112 \\
 & Every step & 1.000 $\pm$ 0.000 & 0.867 $\pm$ 0.070 & 0.184 $\pm$ 0.018 & 0.173 $\pm$ 0.011 \\
\midrule
\multirow{2}{*}{DriftLite}
 & None       & 0.011 $\pm$ 0.005 & 5.973 $\pm$ 1.857 & 0.658 $\pm$ 0.188 & 0.534 $\pm$ 0.136 \\
 & Every step & 1.000 $\pm$ 0.000 & 1.226 $\pm$ 0.117 & 0.124 $\pm$ 0.009 & 0.146 $\pm$ 0.007 \\
\midrule
IPG    & None & -- & \textbf{0.850 $\pm$ 0.054} & 0.013 $\pm$ 0.003 & 0.093 $\pm$ 0.002 \\
IPG-CV & None & -- & \textbf{0.850 $\pm$ 0.047} & \textbf{0.011 $\pm$ 0.002} & \textbf{0.092 $\pm$ 0.002} \\
\bottomrule
\end{tabular}
\end{table}

FKC and DriftLite show improvements, but both are still behind IPG and IPG-CV, which perform as well as they do when using overdamped Langevin.
This indicates that the SVGD-like interactions alone are not enough to replace using the IPG corrective drift.

\subsection{Neural network based corrective drifts} \label{sec:gmmneural}

We briefly explore parameterizing the corrective drift $u_t$ with a neural network.
At each time step $t$, the network weights are warm-started from the previous time step and updated for $50$ iterations by minimizing the objective~\eqref{eq:ipsobjective} without regularization.
We consider approximating $\mathbb{E}_{p_t}[g_t]$ in the objective both with and without the Stein control variate.
Evaluating the objective requires computing $S_{p_t} u_t(x)$ involving $\nabla \cdot u_t(x)$, which we approximate using Hutchinson's trace estimator, where the Jacobian-vector product (JVP) is obtained with \texttt{torch.func.jvp}.

We consider an MLP with 2 hidden layers of size $512$ and GELU activations. 
Additionally, to obtain an interacting drift, we consider a small transformer with no positional encoding that treats each particle position as a token.
The network consists of 2 pre-norm encoder layers with 8 attention heads, a feedforward width of size $4 \times 128$, GELU activations and a linear output layer.
Similar to DriftLite, as the reweighting term may not be driven close to zero here, these are run as weighted particle systems, but without resampling.
We denote the drifts trained with the control variate in the objective using the suffix ``-CV'', and report the metrics in Table~\ref{tab:gmmneural}.

\begin{table}[t]
\centering
\small
\caption{Mean and standard deviation of the metrics using neural network based corrective drifts.}
\label{tab:gmmneural}
\begin{tabular}{@{}llcccc@{}}
\toprule
Method & Resampling & ESS/$N$ $\uparrow$ & Mean error $\downarrow$ & MMD $\downarrow$ & SWD $\downarrow$ \\
\midrule
MLP            & None & 0.982 $\pm$ 0.003 & 0.649 $\pm$ 0.089 & 0.022 $\pm$ 0.002 & 0.089 $\pm$ 0.004 \\
MLP-CV         & None & 0.983 $\pm$ 0.004 & 0.697 $\pm$ 0.035 & 0.019 $\pm$ 0.004 & 0.089 $\pm$ 0.001 \\
Transformer    & None & 0.986 $\pm$ 0.002 & 0.675 $\pm$ 0.033 & 0.020 $\pm$ 0.002 & 0.088 $\pm$ 0.001 \\
Transformer-CV & None & 0.985 $\pm$ 0.001 & 0.687 $\pm$ 0.020 & 0.021 $\pm$ 0.003 & 0.089 $\pm$ 0.002 \\
\bottomrule
\end{tabular}
\end{table}

The neural network based corrective drifts achieve marginally lower mean error and SWD than IPG and IPG-CV, but higher MMD.
Moreover, computing the divergence $\nabla \cdot u_t(x)$ and optimizing the objective with inner iterations at each time step significantly increases the computational cost without a corresponding large gain in performance.

\section{Additional details on image inpainting} \label{sec:imageappendix}

\paragraph{Prior and Reward.} We use the 1-rectified flow model of \citet{liu2022flow}, trained on high-resolution ($256 \times 256$) AFHQ Cat \citep{choi2020stargan} images, as our prior $q_1(x)$.
For the inpainting task, observations are obtained by either applying a square mask with side length $0.3 \times$ the image size at the center of the image, or masking the right half of the image.
In both cases, we add Gaussian noise with $\sigma_\text{obs}^2 = 0.05$.

All methods are run by uniformly discretizing time $t \in [0,1]$ into 200 steps.
For the Langevin noise schedule, we use $\sigma_t = 1-t$, since the score $\nabla \log q_t(x)$, and in turn $\nabla \log p_t(x)$, blows up as $t$ approaches $1$.
In DriftLite, the divergences required to evaluate $S_{p_t} u_t(x)$ are approximated with Hutchinson's trace estimator using the central difference method
\begin{equation}
    \nabla \cdot f(x) \approx \frac{1}{2 \delta}\langle \epsilon, f(x + \delta \epsilon) - f(x - \delta \epsilon) \rangle,
\end{equation}
where $\delta = 0.01$ and $\epsilon \sim \mathcal{N}(0, I)$. For IPG and IPG-CV, we use $\lambda=10^{-1}$ which keeps the linear solve well-conditioned, and verify that ESS/$N \geq 0.999$ if weighted particles are used.

\paragraph{Metrics.} PSNR, SSIM and LPIPS are computed against the ground truth image using their standard implementation in TorchMetrics. To measure diversity among the generated samples $\{x^i\}_{i=1}^N$ per test image, we report the sample covariance trace
\begin{equation}
    \text{Cov. trace} = \frac{1}{N-1} \sum_{i=1}^N \| x^i - \bar{x} \|_2^2, \quad \bar{x} = \frac{1}{N} \sum_{i=1}^N x^i,
\end{equation}
where the per-pixel values are in $[-1, 1]$, and the mean pairwise LPIPS,
\begin{equation}
    \text{LPIPS div.} = \frac{2}{N(N-1)} \sum_{i < j} \text{LPIPS}(x^i, x^j).
\end{equation}
For both measures, values close to zero indicate particle collapse.

\begin{figure}[t]
  \centering
  \begin{subfigure}[t]{0.8\textwidth}
    \centering
    \includegraphics[width=\textwidth]{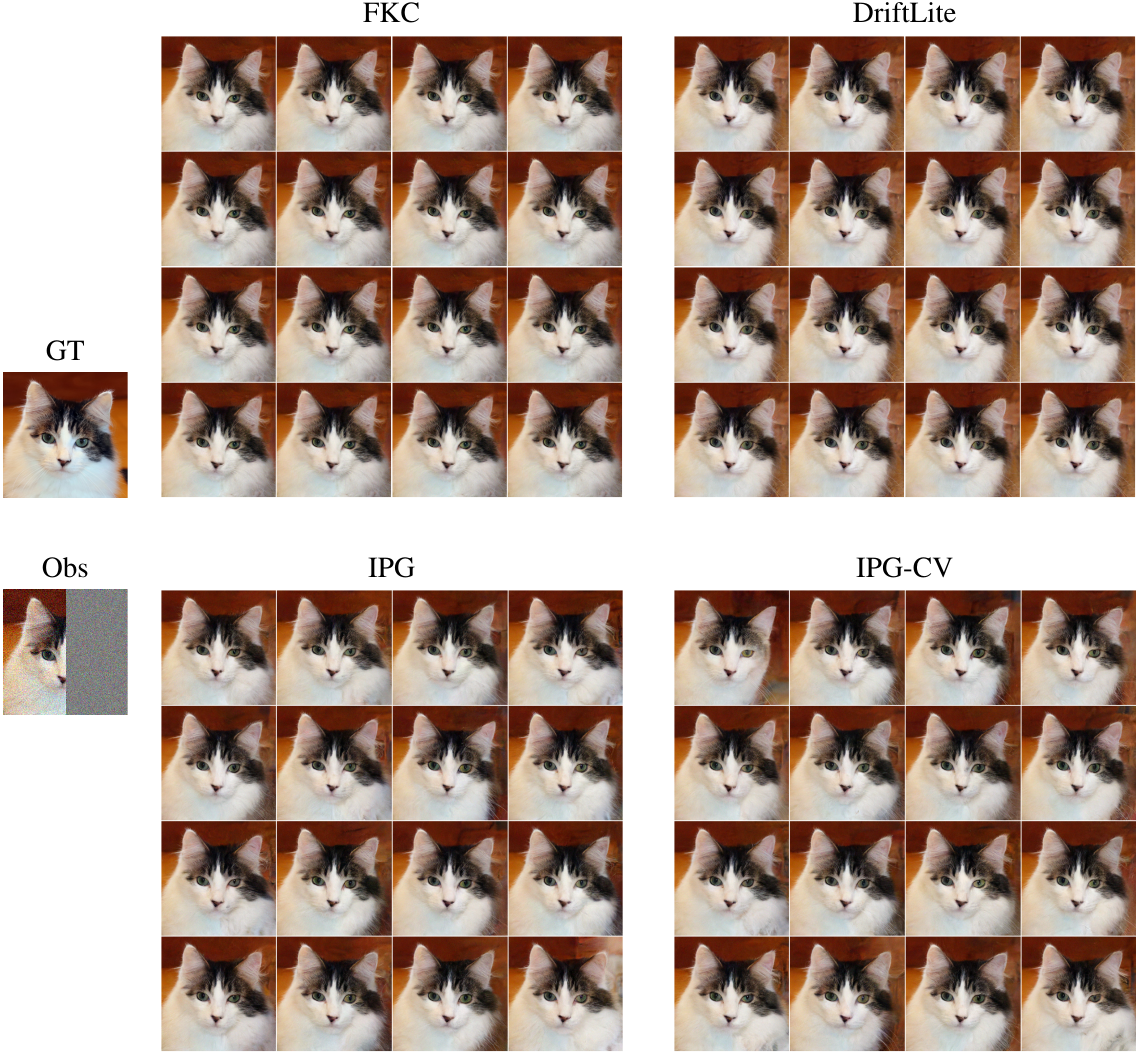}
  \end{subfigure}
  \vfill
  \begin{subfigure}[b]{0.8\textwidth}
      \centering
      \includegraphics[width=\textwidth]{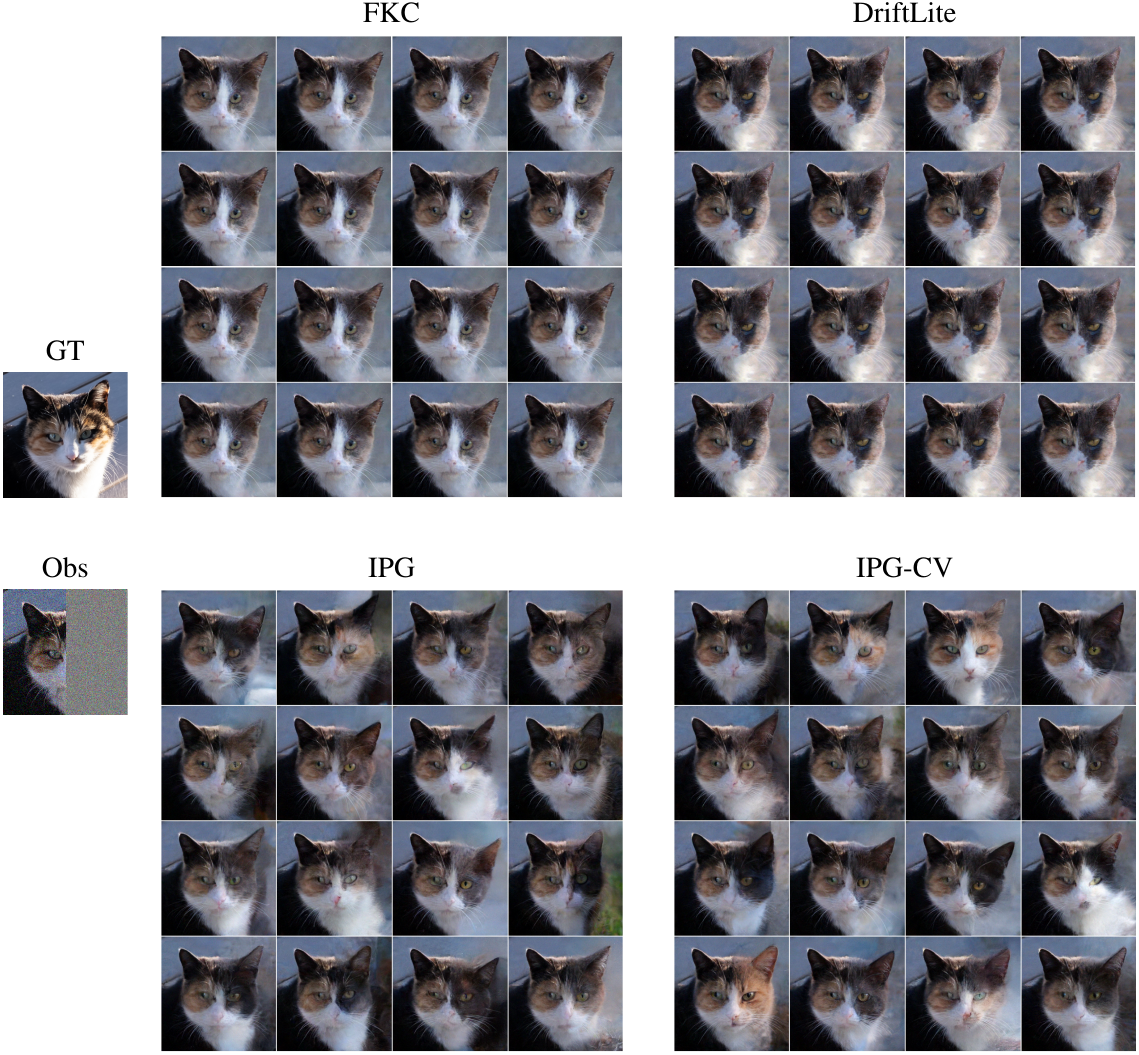}
  \end{subfigure}
  \caption{Samples produced by the methods along with the ground truth (GT) and observation (Obs) for the half-mask inpainting task.}
  \label{fig:outpaint_full}
\end{figure}

\begin{figure}[t]
  \centering
  \begin{subfigure}[t]{0.8\textwidth}
    \centering
    \includegraphics[width=\textwidth]{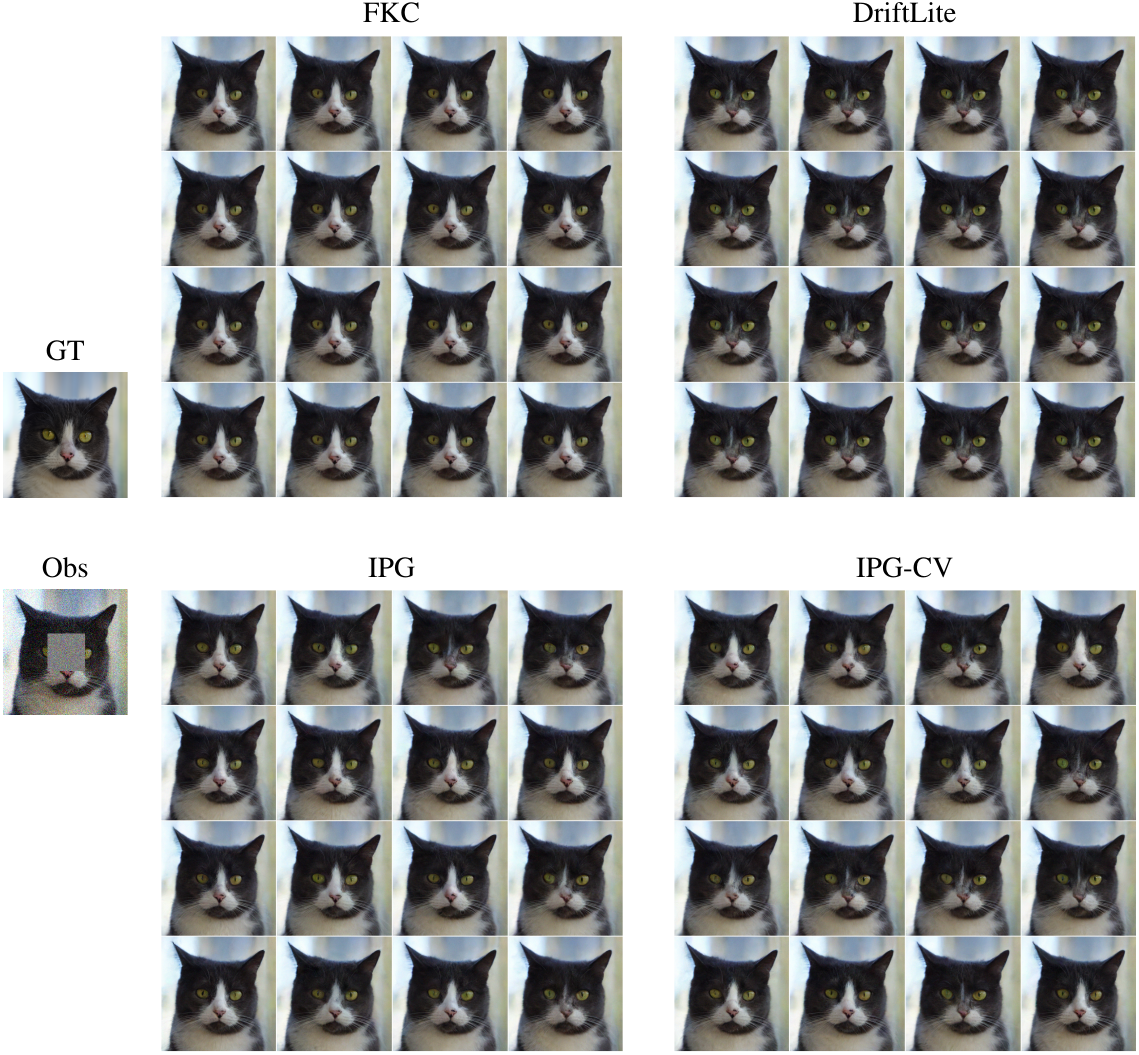}
  \end{subfigure}
  \vfill
  \begin{subfigure}[b]{0.8\textwidth}
      \centering
      \includegraphics[width=\textwidth]{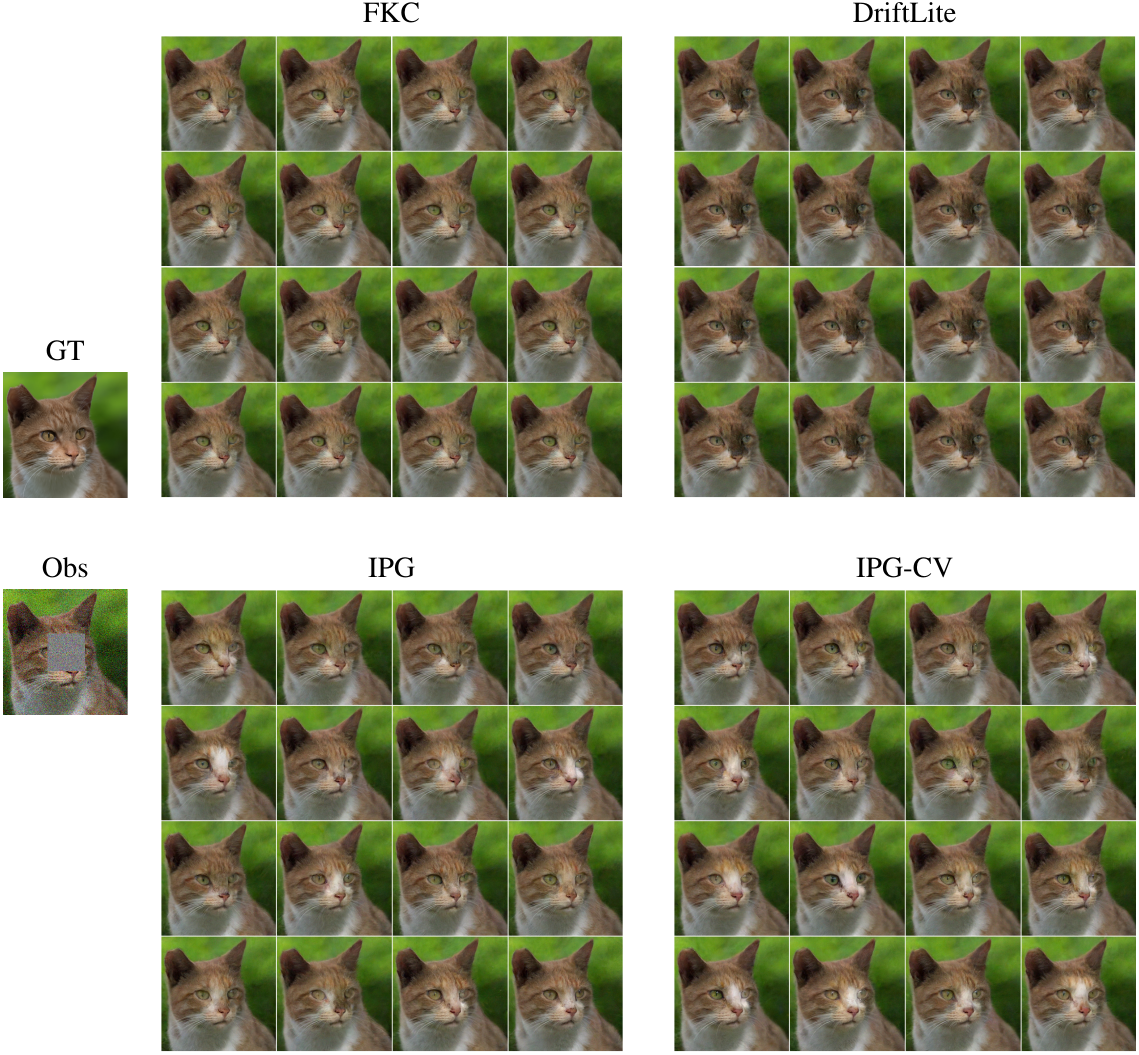}
  \end{subfigure}
  \caption{Samples produced by the methods along with the ground truth (GT) and observation (Obs) for the box-mask inpainting task.}
  \label{fig:inpaint_full}
\end{figure}

We visualize the samples produced by different methods for inpainting in Figures~\ref{fig:outpaint_full} and \ref{fig:inpaint_full}.
While IPG and IPG-CV produce diverse samples from the posterior, both FKC and DriftLite suffer from severe particle collapse.

\section{Additional details on protein backbone structure inference} \label{sec:proteinappendix}

\paragraph{Prior and Reward.} We use Proteina \citep{geffner2025proteina}, specifically the $60$M parameter flow-based model that generates protein backbone structures, as our prior $q_1(x)$.
It is trained on a subset of AFDB \citep{varadi2022alphafold} which includes synthetic structures up to $256$ residues in length, and predicts only the $\alpha$-carbon position in each residue.
For the observation, we take $K$ pairwise distances between uniformly sampled distinct pairs of residues, with $K$ as $3\%$ of all available pairs, and add Gaussian noise with $\sigma_\text{obs} = 1.0$ \AA. 

\paragraph{Metrics.} To ensure that the generated structures do not deviate substantially from the observation, we measure the RMSD between the pairwise distances in the samples and those in the observation $y$ ($\text{RMSD}_\text{obs}$).
We also measure the RMSD between the generated samples and the ground truth after rotational alignment using the Kabsch algorithm ($\text{RMSD}_\text{gt}$). 
Following common practice, the quality of generated protein structures is assessed with the following procedure.
\begin{enumerate}
    \item For each generated structure, $8$ amino-acid sequences are predicted using the inverse folding model ProteinMPNN \citep{dauparas2022robust}.
    \item Each of these sequences is refolded using ESMFold \citep{lin2023evolutionary}, and the RMSD between the refolded structure and the original generated structure is measured.
    \item If the lowest such RMSD, defined as the self-consistency RMSD, is below $2$ \AA, then the structure is considered designable.
\end{enumerate}
We report the fraction of test cases in which a method produces at least one designable structure as ``Any designable''.

The methods are run by uniformly discretizing time $t \in [0,1]$ into 500 steps.
As in the image inpainting task, we use the Langevin noise schedule $\sigma_t = 1-t$, and compute the divergences required for DriftLite with Hutchinson's trace estimator using the central difference method.
For IPG and IPG-CV, we use $\lambda=10^{-2}$ which keeps the linear solve well-conditioned and yields ESS/$N = 0.999$ with weighted particles.
We note that Proteina, when used with the generative ODE, produces designable samples at a rate below $20\%$.
\citet{geffner2025proteina} suggest using a heuristic generative SDE with noise level $\gamma < 1$ to improve designability at the cost of diversity,
\begin{equation} \label{eq:proteina}
    \diff X_t = (v_t(X_t) + \sigma_t \nabla \log q_t(X_t)) \diff t + \sqrt{2\sigma_t \gamma} \diff W_t.
\end{equation}
However, for $\gamma \neq 1$, the marginal densities of this SDE are not $q_t(x)$, and the score in the Langevin term within our framework can no longer be estimated from $v_t(x)$.
While equation~\eqref{eq:proteina} can be used for the particle updates by adjusting the reweighting term accordingly, we leave this for future work and do not use it in our experiments for simplicity.

\paragraph{Test cases.} We collect $15$ structures from the Protein Data Bank that were deposited after the release date of the model weights, with lengths between $100$ and $250$ residues.
All structures were determined using X-ray diffraction, with the exception of \texttt{9ZKD}, which was determined using electron microscopy, and all were deposited at a resolution at or below $2.5$ \AA.
Further details on the individual structures are provided in Table~\ref{tab:proteintests}.

\begin{figure}[t]
\begin{center}
\includegraphics[width=0.6\linewidth]{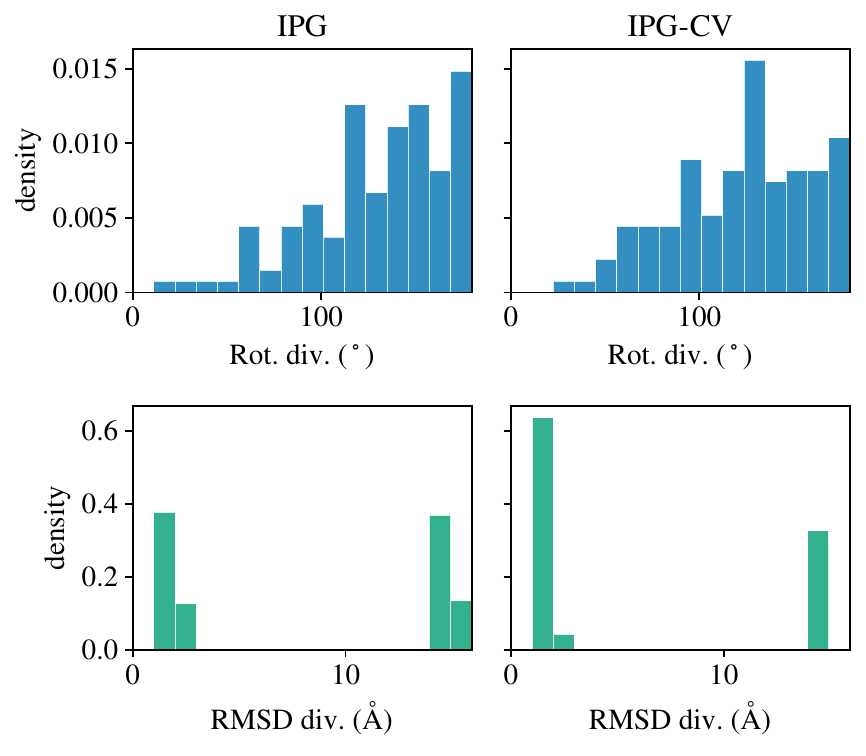}
\end{center}
\caption{Distribution of pairwise diversity metrics for IPG and IPG-CV on \texttt{30JI}. The bimodal RMSD div. indicates that structures close to reflections of the ground truth are occasionally sampled due to the imperfect generative prior.}
\label{fig:protein_div}
\end{figure}

\begin{figure}[t]
\begin{center}
\includegraphics[width=0.8\linewidth]{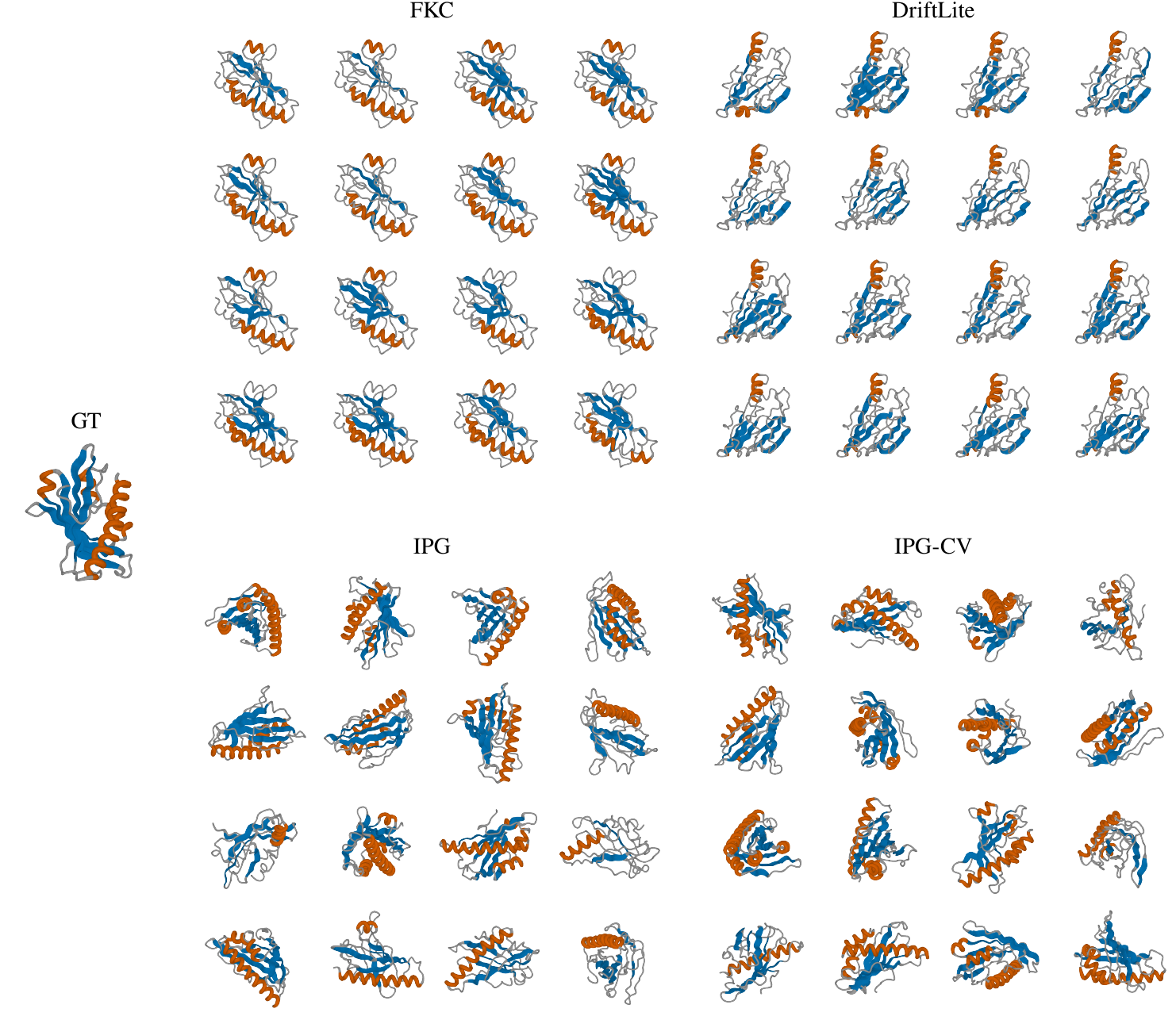}
\end{center}
\caption{Samples produced from observing random inter-residue distances for protein \texttt{30JI}, visualized without Kabsch alignment to the ground truth (GT).}
\label{fig:protein_app1}
\end{figure}

\begin{figure}[t]
\begin{center}
\includegraphics[width=0.8\linewidth]{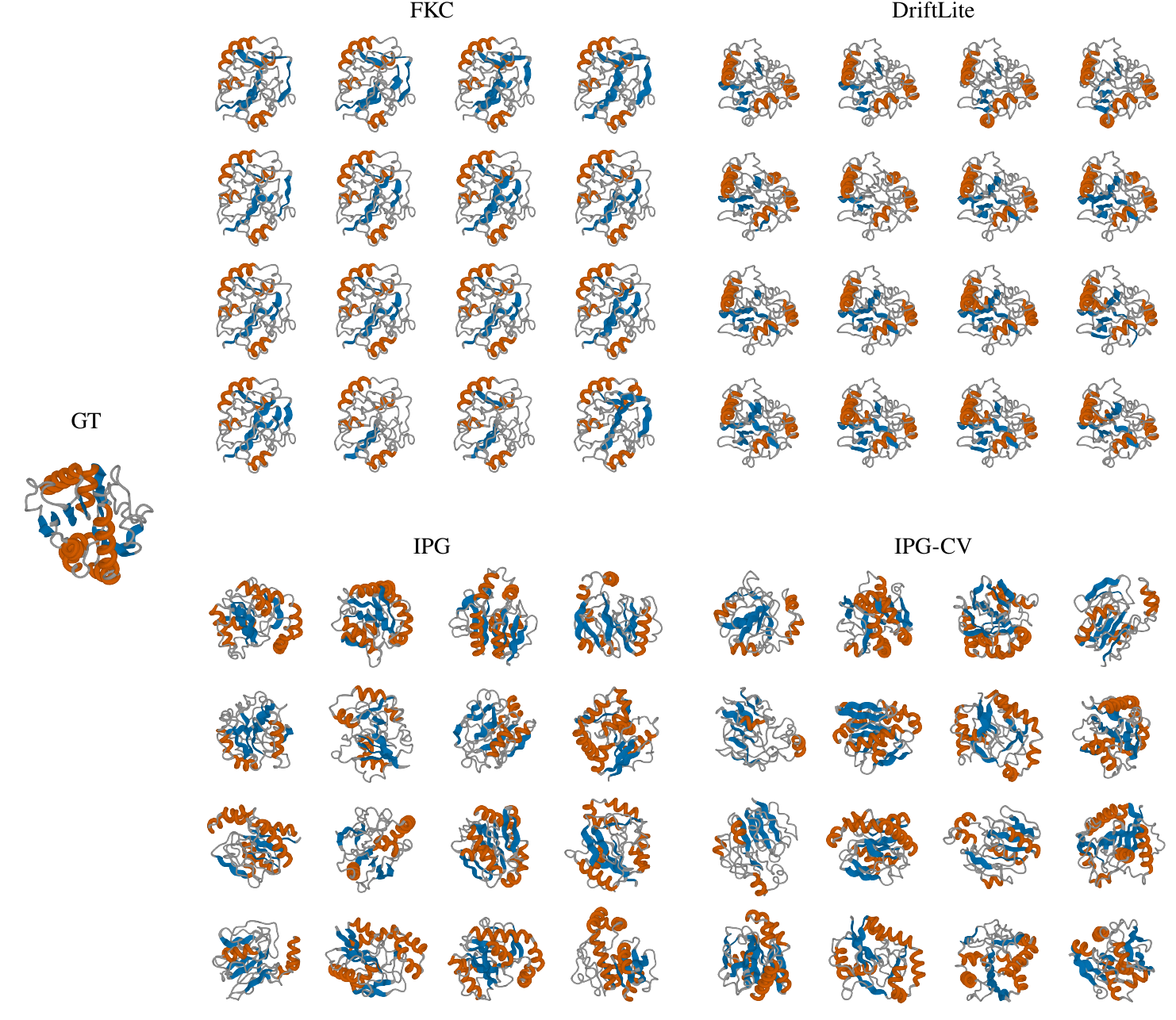}
\end{center}
\caption{Samples produced from observing random inter-residue distances for protein \texttt{9XFS}, visualized without Kabsch alignment to the ground truth (GT).}
\label{fig:protein_app2}
\end{figure}

Samples produced by different methods are visualized in Figures~\ref{fig:protein_app1} and \ref{fig:protein_app2}.
As Proteina predicts only the $\alpha$-carbon positions, secondary structures are assigned based on the P-SEA algorithm from the \texttt{biotite} package \citep{kunzmann2018biotite}.
Samples are intentionally not rotation aligned with the ground truth structure to highlight the particle collapse in FKC and DriftLite, while IPG and IPG-CV produce random orientations, as expected.
Histograms of pairwise diversity metrics for IPG and IPG-CV on one test case (\texttt{30JI}) are shown in Figure~\ref{fig:protein_div}.
The bimodal distribution of RMSD div. indicates that the methods produce some undesignable structures close to reflected versions of the ground truth. This is to be expected when the designability of the Proteina prior is below $20\%$ as discussed above.

\begin{table}[t]
\centering
\small
\caption{Target proteins used for the structure inference task. In some cases, the terminal residues are not resolved, resulting in a lower modeled residue count than the deposited count.}
\label{tab:proteintests}
\begin{tabular}{@{}lccc@{}}
\toprule
 & \multicolumn{2}{c}{Residue count} & \\
\cmidrule(lr){2-3}
PDB ID & Modeled & Deposited & Resolution (\AA) \\
\midrule
12FG & 166 & 171 & 1.43 \\
28HS & 138 & 142 & 1.10 \\
28RZ & 186 & 186 & 2.50 \\
30JI & 173 & 173 & 1.75 \\
30ZU & 223 & 246 & 1.05 \\
32HC & 190 & 197 & 1.40 \\
38LJ & 210 & 216 & 1.40 \\
9SM9 & 129 & 129 & 1.96 \\
9WGR & 134 & 145 & 1.53 \\
9WRY & 205 & 211 & 2.02 \\
9XFS & 230 & 241 & 2.20 \\
9YKH & 165 & 170 & 1.02 \\
9YZ6 & 128 & 128 & 2.47 \\
9ZKD & 208 & 208 & 2.32 \\
9ZU4 & 103 & 129 & 1.05 \\
\bottomrule
\end{tabular}
\end{table}

\end{document}